\documentclass[lettersize,journal,twoside]{IEEEtran}

\usepackage[hyphens]{url}
\usepackage[pdfstartview=XYZ,
bookmarks=true,
colorlinks=true,
linkcolor=blue,
urlcolor=blue,
citecolor=blue,
pdftex,
bookmarks=true,
linktocpage=true, 
hyperindex=true
]{hyperref}

\usepackage{orcidlink}

\usepackage{amsmath,amsfonts}

\usepackage{algorithm}
\usepackage{algpseudocode}

\usepackage[utf8]{inputenc}

\usepackage{array}
\usepackage[caption=false,font=normalsize,labelfont=sf,textfont=sf]{subfig}
\usepackage{textcomp}
\usepackage{stfloats}
\usepackage{verbatim}
\usepackage{graphicx}
\usepackage{cite}
\usepackage{multirow}
\usepackage{todonotes}
\usepackage{xcolor}
\usepackage{comment}

\begin{document}

\title{An All-digital 8.6-nJ/Frame 65-nm Tsetlin Machine Image Classification Accelerator}

\author{
Svein Anders Tunheim$^{\orcidlink{0000-0001-7947-8485}}$,~\IEEEmembership{Senior Member,~IEEE,}
Yujin Zheng$^{\orcidlink{0000-0001-8810-8857}}$, 
Lei Jiao$^{\orcidlink{0000-0002-7115-6489}}$,
~\IEEEmembership{Senior Member,~IEEE,} \\
Rishad Shafik$^{\orcidlink{0000-0001-5444-537X}}$,~\IEEEmembership{Senior Member,~IEEE,} 
Alex Yakovlev$^{\orcidlink{0000-0003-0826-9330}}$,~\IEEEmembership{Fellow,~IEEE}, and 
Ole-Christoffer Granmo$^{\orcidlink{0000-0002-7287-030X}}$


\thanks{This work was funded by the University of Agder, Norway, and the Research Council of Norway under Grant 347712. 

Svein Anders Tunheim, Lei Jiao and Ole-Christoffer Granmo are with the Centre for Artificial Intelligence Research (CAIR), The University of Agder, Jon Lilletuns vei~9, 4879 Grimstad, Norway.

Yujin Zheng, Rishad Shafik and Alex Yakovlev are with the Microsystems Group, School of Engineering, Newcastle University, Newcastle upon Tyne NE1~7RU,~UK.
\vspace{50pt}
}

}



\markboth{Accepted for publication in IEEE Transactions on Circuits and Systems---I: Regular Papers}
{Tunheim
\MakeLowercase{\textit{et al.}}: An All-digital 8.6-\MakeLowercase{n}J/Frame 65-\MakeLowercase{nm} Tsetlin Machine Image Classification Accelerator}

\IEEEpubid{
\textit{Copyright 2025 IEEE. Personal use is permitted, but republication/redistribution requires IEEE permission.}} 





\maketitle

\begin{abstract}
We present an all-digital programmable machine learning accelerator chip for image classification, underpinning on the Tsetlin machine (TM) principles. The TM is an emerging machine learning algorithm founded on propositional logic, utilizing sub-pattern recognition expressions called clauses. The accelerator implements the coalesced TM version with convolution, and classifies booleanized images of~28$\times$28 pixels with~10 categories. A configuration with~128~clauses is used in a highly parallel architecture. Fast clause evaluation is achieved by keeping all clause weights and Tsetlin automata (TA) action signals in registers. The chip is implemented in a~65~nm low-leakage CMOS technology, and occupies an active area of~2.7~mm$^2$. At a clock frequency of~27.8~MHz, the accelerator achieves~60.3~k classifications per second, and consumes~8.6~nJ per classification. This demonstrates the energy-efficiency of the TM, which was the main motivation for developing this chip. The latency for classifying a single image is~25.4~$\mu$s which includes system timing overhead. The accelerator achieves 97.42\%,~84.54\% and~82.55\% test accuracies for the datasets MNIST, Fashion-MNIST and Kuzushiji-MNIST, respectively, matching the TM software models.
\end{abstract}


\section{Introduction}\label{PaperD_Section: Introduction}

\IEEEPARstart{H}{ardware} (HW) accelerators are commonly integrated in modern electronic systems to relieve the burden on system processors from extensive application-specific workloads. Performance and energy-efficiency within a smaller form factor are key considerations for accelerator designs, particularly for Internet-of-Things (IoT), where the compute and memory resources of edge nodes typically are limited. In addition, such nodes are often battery-operated, and it is critical to minimize the overall energy footprints for longer battery operational lifetime. 

Image classification is a key machine learning (ML) application, with usage scenarios across a wide range of IoT applications~\cite{Thesis_maheepala2020low}, such as industrial machine-vision, agricultural monitoring, unmanned aerial vehicles, room occupancy detection for building control systems, attention detection and gesture recognition. 

Current methods for accelerating image classification in embedded systems, are predominantly based on convolutional neural networks~(CNNs)~\cite{Thesis_8662396LeCunISSCC2019, Thesis_8114708_Sze_survey_IEE_Proceedings}. CNNs comprise many layers of multiply-and-accumulate (MAC) operations, each with their associated memory resources. As such, they require {\color{black}a} large number of micro-architectural operations for each classification and have significant energy footprints~\cite{Thesis_8114708_Sze_survey_IEE_Proceedings}. The complexity and energy consumption can be reduced by compromising the data precision~\cite{Thesis_9063049_Jeff_dean}. Binary and ternary activations and weights have been applied in several integrated circuit~(IC) solutions that have shown good test accuracy with very low power consumption~\cite{Thesis_BankmanBNNmixedsignal, Thesis_AlwaysOnBNNDiMauro, Thesis_KnagPhilC2021A6AB, Thesis_YodaNN7878541, Thesis_10034979_Yang_TNN_MNIST_May2023}. 

\IEEEpubidadjcol

Unlike the numerical models based on neural networks mentioned above, the Tsetlin machine~\cite{Thesis_OrigTM} is an ML algorithm founded on propositional logic. Parallel conjunctive \textit{clauses} are defined by independent learning automata, and each clause represents a logical sub-pattern. The TM features a single-layer structure with highly interpretable outputs, making it well-suited for application areas such as biomedical, cybernetics, IoT, and law. Furthermore, because its primary operations are of Boolean type, TM solutions are {\color{black}HW-friendly}. The generation of a TM's output relies on summing and comparing class sums, with minimal or no need for energy-intensive multiplication operations, even during training. This makes TMs ideal for low-energy application-specific integrated circuit (ASIC) implementations~\cite{Thesis_Wheeldon1}. Additionally, due to their small resource footprint, TMs facilitate hardware solutions that support on-device training and continuous learning~\cite{Thesis_10812055_TCSI_ConvCoTM28x28_FPGA}.

The original TM version~\cite{Thesis_OrigTM} is called the \textit{vanilla TM}. For image classification, improved test accuracy is achieved by the \textit{convolutional TM~(CTM)}~\cite{Thesis_CTM}. The CTM achieved a peak test accuracy of 99.4\% on the MNIST dataset~\cite{Thesis_lecun-mnisthandwrittendigit-2010}. For the Fashion-MNIST (FMNIST)~\cite{Thesis_Fashion-MNIST_dataset} and Kuzushiji-MNIST (KMNIST)~\cite{Thesis_Kuzushiji-MNIST_dataset} datasets, the peak test accuracies achieved were 91.5\% and 96.31\% respectively. Ensemble TM approaches have shown promising accuracies for more complex image datasets, such as \mbox{CIFAR-10} and \mbox{CIFAR-100}, which are currently being investigated. In the case of \mbox{CIFAR-10}, the 
approach denoted \textit{TM Composites}~\cite{Thesis_granmo2023tmcomposites} has demonstrated a test accuracy of~82.8\%~\cite{Thesis_grønningsæter2024optimizedtoolboxadvancedimage}.

In this paper, we present an inference accelerator ASIC based on the \textit{coalesced TM}~(CoTM)~\cite{Thesis_CoalescedTM}. The CoTM utilizes a single clause pool, in contrast to the vanilla TM which uses one clause pool per class. As we operate the CoTM with convolution, we here denote it as a \textit{convolutional CoTM}~(\textit{ConvCoTM}). For configurations with few clauses, the ConvCoTM achieves better accuracy than the vanilla CTM, making it attractive for resource-constrained solutions.

Until now, only a single ASIC based on the TM has been reported~\cite{Thesis_Wheeldon1}{\color{black}, and it} operated on a simple dataset. We wanted to make an an accelerator that operated on a well-known dataset (MNIST) and to demonstrate that an all-digital TM-based ASIC could achieve energy-efficiency at the same level as analog in-memory-computing (IMC)  solutions~\cite{Thesis_10902457_3p32nJ_per_Frame, Thesis_10058600_Yejun_Ko_SNN_TCSI_II, Thesis_10034979_Yang_TNN_MNIST_May2023}.  

{\color{black}The ConvCoTM FPGA solution in~\cite{Thesis_10812055_TCSI_ConvCoTM28x28_FPGA} demonstrated inference and full-on device training. However, ultra-low-power (ULP) operation could not be achieved with the FPGA. As full on-device TM-based training had been proved feasible in~\cite{Thesis_10812055_TCSI_ConvCoTM28x28_FPGA}, we wanted to design an \textit{inference-only} ASIC solution, optimized for low-power operation. 

The ASIC accelerator's inference core is built on modules included in~\cite{Thesis_10812055_TCSI_ConvCoTM28x28_FPGA}, with several enhancements} as described in section~\ref{PaperD_Section: Design Overview}. In particular, there is an updated clause logic circuit with a novel feedback that reduces switching activity. There is also new functionality for loading a pre-trained model. Most important is the 
optimization of the inference part to achieve ULP operation. 

Although MNIST is a simple dataset, it is still widely used for benchmarking of ULP image classification ASICs~\cite{Thesis_10902457_3p32nJ_per_Frame, Thesis_10058600_Yejun_Ko_SNN_TCSI_II, Thesis_10034979_Yang_TNN_MNIST_May2023}. We therefore consider the MNIST dataset suitable for demonstrating the HW-friendliness and power-efficiency of a ConvCoTM-based accelerator, which was the main motivation for the design and implementation of this ASIC. The main contributions of our paper are:

\begin{itemize}
    
    \item The accelerator is the first reported manufactured ASIC based on the CoTM. It is also the first TM ASIC where convolution is utilized. 
    
    \item The accelerator is a fully digital, synchronous design, compatible with standard digital workflows. It operates with an energy per classification (EPC) of~8.6~nJ, which is the lowest reported for a fully digital solution running on the same dataset and with comparable test accuracy. Overall, it ranks as the second most energy-efficient solution. Other comparable solutions are of analog and mixed-signal type, utilizing IMC~\cite{Thesis_10902457_3p32nJ_per_Frame, Thesis_10058600_Yejun_Ko_SNN_TCSI_II, Thesis_10034979_Yang_TNN_MNIST_May2023}.
  
    \item The accelerator has low latency, essential for real-time operation in low-power systems with strict requirements.
    
    \item For TMs operating with convolution, we introduce a novel clause logic circuit with feedback, which reduces switching of {\color{black} a} clause's combinational output.

    \item We describe an envisaged scaled-up ASIC that operates on the CIFAR-10 dataset, and provide estimates of its performance.
    
\end{itemize}

The remainder of the paper is organized as follows: Section~\ref{PaperD_Section: Related work} summarizes related work. Section~\ref{PaperD_Section: TM and CoTM Background} covers TM and CoTM background, including convolution operation. Section~\ref{PaperD_Section: Design Overview} details the ASIC's architecture and building blocks{\color{black}, and m}easurement results are presented in Section~\ref{PaperD_Section: Implementation and Measurement Results}. {\color{black} In Section~\ref{PaperD_Section: Possible extensions} we discuss possible extensions and modifications of the design. The ConvCoTM accelerator's performance is discussed in Section~\ref{PaperD_Section: Discussions}, and Section~\ref{PaperD_Conclusion} concludes the paper.}
\section{Related Work}\label{PaperD_Section: Related work}

In this section, we describe state-of-the-art works on low-power inference accelerators for image classification, as well as some reported TM-based HW solutions. 

For low-power neural networks, digital {\color{black}b}inary {\color{black}n}eural {\color{black}n}etworks (BNNs) are widely utilized. A BNN-based accelerator IC module is detailed in ~\cite{Thesis_XNORNeuralEngine}. This module was later included in a complete system-on-chip (SoC)~\cite{Thesis_AlwaysOnBNNDiMauro}. When operating on the CIFAR-10~\cite{Thesis_Krizhevsky09learningmultiple} dataset, the SoC achieves~15.4~inferences/s and has a peak power envelope of~$674~\mu$W. A~22~nm fully depleted silicon on insulator (FD-SOI) technology was used for the IC implementation.

In~\cite{Thesis_KnagPhilC2021A6AB} a digital~10~nm FinFET CMOS solution is reported. Utilizing a BNN algorithm, it achieves a peak energy efficiency of~617~{\color{black}Tera operations per second per watt (TOPS/W)}. It has a power consumption of~5.6~mW and obtains an accuracy of 86\% when operating on the CIFAR-10 dataset.



IMC solutions can provide very high energy-efficiency, as the bottleneck related to memory access is relieved. IMC has especially been utilized for analog/mixed-signal ML accelerator designs. An IMC mixed-signal 28~nm CMOS BNN processor is described in~\cite{Thesis_BankmanBNNmixedsignal}. This accelerator achieves an EPC of~3.8~$\mu$J when operating on \mbox{CIFAR-10}. The classification rate is~237 frames per second~(FPS) with a power consumption of~0.9~mW.

In \cite{Thesis_10034979_Yang_TNN_MNIST_May2023} a charge-domain IMC ternary neural network (TNN) chip is described. The main advantage of the ternary approach is a reduction of the required operations per classification with~3.9$\times$ compared to a BNN model. The accelerator classifies MNIST images at a rate of~549~FPS, consuming~96~$\mu$W, {\color{black}which} corresponds to an EPC of~0.18~$\mu$J. The test accuracy achieved is~97.1\%. 

The accelerator chip that achieves the lowest reported EPC for MNIST classification, is reported in~\cite{Thesis_10902457_3p32nJ_per_Frame}. It is a CNN utilizing an analog IMC architecture with time domain signal processing. It is implemented in~28nm CMOS and achieves an EPC of only~3.32~nJ. 

Neuromorphic approaches, utilizing spiking neural networks~(SNNs) have achieved very low energy consumption.  
In~\cite{Thesis_10058600_Yejun_Ko_SNN_TCSI_II} a mixed-signal neuromorphic SNN IC is reported, which has been manufactured in a~65~nm CMOS technology. It achieves a test accuracy on MNIST of~95.35\%, with a power consumption of~0.517~mW and an EPC of~12.92~nJ. Simulation results of a~28~nm CMOS charge-domain computing SNN solution are reported in~\cite{Thesis_10396030_Yuchao_Zhang_SNN_ASICON}. The accelerator achieves an EPC of~15.09~nJ, a latency of~0.46~$\mu$s and a test accuracy on MNIST of~94.81\%. In~\cite{Thesis_10388821_Wenbing_Fang_SNN_BioCAS}, simulation results of a clock-free, event-driven, mixed-signal~28~nm IMC SNN are described. On MNIST it achieves a test accuracy of~96.92\% and~1.38~nJ EPC. {\color{black}A time-domain IMC mixed-signal SNN is reported in~\cite{Thesis_10734367_SNN_701p7_TOPSperW}. It is implemented in a 65~nm CMOS technology and has an energy efficiency of 701.7~TOPS/W. On the CIFAR-10 dataset it achieves a test accuracy of 91.3\%. 

In~\cite{Thesis_ACIM_2025_JSSC_10689660} an analog IMC accelerator suited for CNNs as well as for Transformers is presented. It achieves 818–4094~TOPS/W, and CIFAR-10 test accuracies of~91.7\% and~95.8\% for CNN (ResNet20) and Transformer (ViT-S) respectively. The technology used is~65~nm CMOS.
}

For TM-based HW solutions, the first reported ASIC is described in~\cite{Thesis_Wheeldon1}. It utilizes the vanilla TM architecture and operates on a 3-class machine learning (ML) problem. The chip is implemented in a~65~nm CMOS technology and supports both training and inference.  In~\cite{Thesis_MICPROTUNHEIM2023104949}, the first TM-based HW accelerator with convolution is reported. It is an FPGA solution which operates on a~2-class pattern recognition problem (two-dimensional noisy XOR~\cite{Thesis_CTM}) in~4$\times$4 Boolean images, and employs a~2$\times$2 convolution window. Full on-device training is implemented. 

The FPGA solution in~\cite{Thesis_10455016_Prajwal} is based on the vanilla TM and utilizes the sparsity of typical TM models, in combination with sequential operation. It represents a highly compact solution when it comes to HW resource usage. 

In~\cite{Thesis_10812055_TCSI_ConvCoTM28x28_FPGA} 
the first reported ConvCoTM-based HW solution is presented, of which the inference part is the basis for our ASIC{\color{black},} as described in the Introduction section. It is an FPGA solution that classifies~$28 \times 28$ images with~10 categories, and includes full on-device training. For the MNIST dataset a test accuracy of~97.6\% is achieved, with a classification rate of~134~k samples per second, and an EPC of~13.3~$\mu$J. The source code for this design is available at~\cite{Thesis_GitHub_FPGA_repository}. 

{\color{black}The FPGA-based accelerator in~\cite{Thesis_Dynamic_TM_FPGA_10990173} is a flexible solution that supports on-device training for the vanilla TM and the CoTM. It requires less HW resources and provides greater flexibility compared to~\cite{Thesis_10812055_TCSI_ConvCoTM28x28_FPGA}, at the expense of higher latency.}

In~\cite{Thesis_9474126_asynch, Thesis_9565438_asynch, Thesis_10666312_asynch} the design and simulation of low-latency asynchronous TM solutions for inference and training are presented. A mixed-signal IMC TM solution is described in~\cite{Thesis_ghazal2023imbue} with a simulated EPC for MNIST of~13.9~nJ. It is an inference solution with a Boolean-to-current architecture, and resistive RAM (ReRAM) transistor cells are employed to eliminate the need for energy-hungry analog-to-digital and digital-to-analog conversions. Another IMC TM design concept based on Y-flash cells, is reported in~\cite{Thesis_ghazal2024inmemorylearningautomataarchitecture}. 
\section{TM and CoTM Background} \label{PaperD_Section: TM and CoTM Background}

The basic concepts of a vanilla TM~\cite{Thesis_OrigTM} are described here, followed by a description of general CoTM inference~\cite{Thesis_CoalescedTM}. Later, we explain how to apply convolution to the CoTM. As the emphasis in this paper is on \textit{inference} solutions, we do not include details about CoTM/ConvCoTM training, except for a brief description of the Tsetlin automaton~(TA), which is the core learning element in a~TM. TM~training is covered in detail in the original TM, CTM and CoTM references~\cite{Thesis_OrigTM, Thesis_CTM, Thesis_CoalescedTM}. A compact description of ConvCoTM training can be found in \cite{Thesis_10812055_TCSI_ConvCoTM28x28_FPGA}. 
\vspace{10pt}
\subsection{General TM Concepts}\label{subsection General TM Concepts}

{\color{black}The input feature vector, $X$, to a TM consists of $o$ Boolean variables, 
$X=[x_0, x_1, \dots, x_{o-1}]$.} A new input vector named the \textit{literals}, \textbf{\textit{L}}, is generated by appending the negated variables to the input, as shown in Eq.~(\ref{PaperD_Equation: Literals}). Thus, there are in total~$2o$ {\color{black}single-bit} literals.
\begin{multline}
    \label{PaperD_Equation: Literals}
    L=[l_0, l_1, \dots, l_{2o-1}] \\=[x_0, \dots, x_{o-1}, \neg x_0, \dots, \neg x_{o-1}]
\end{multline}

A TA of two-action type \cite{Thesis_Tsetlin1961} will either \textit{include} or \textit{exclude} a literal in a conjunctive \textit{clause}, depending on its state. {\color{black}Fig.~\ref{PaperD_Figure_TA} shows the structure of a single TA with $2N$ states. In HW, a TA is typically implemented as a binary up/down counter, and the inverted version of its most significant bit (MSB) is used as the \textit{TA action} signal (active high).

\begin{figure}[ht!]
\centering
\includegraphics[width=0.50\textwidth]{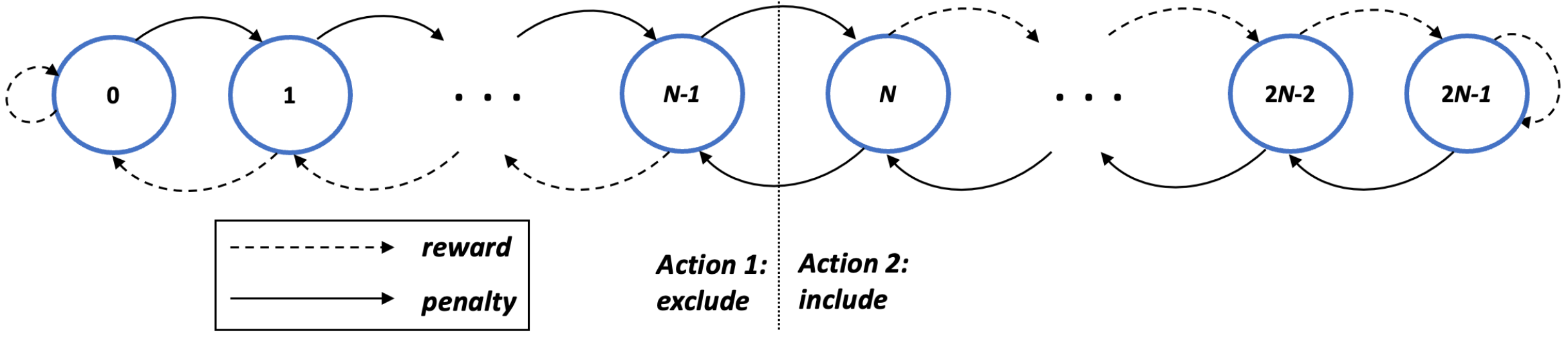}
\caption{{\color{black}A Tsetlin Automaton (TA) for two-action environments.}} 
\label{PaperD_Figure_TA}
\end{figure}
}

One team of TAs is applied per clause, and the different TA actions are obtained during the training process, which involves feedback mechanisms to each TA~\cite{Thesis_OrigTM}. For an inference-only solution, as reported in this paper, only the \textit{TA action signals} from the trained model are needed, not the complete TAs. 

In a TM, the number of clauses, $n$, is a user specified integer. The output of a single clause, $c_j$, where $j \in \{0,\ldots, n-1 \}$, is given by
\begin{equation} \label{PaperD_Equation: eqClauseJ}
    c_j=\bigwedge_{k\in I_j}l_k, 
\end{equation} 
where $l_k$ is the literal with index \textit{k}. $k$ belongs to {\color{black}the set} $ I_j \subseteq \{0, \dots, 2o - 1\}${\color{black},}
{\color{black} which is obtained from training, or from a pre-trained model. $I_j$ consists of the} indices of the literals in clause $j$ for which the corresponding TAs take the action \textit{include}.

\subsection{CoTM} \label{PaperD_subsection: CoTM inference principle}

A vanilla TM employs one clause pool per class, and the clauses of each TM are grouped in two: positive and negative. In contrast, a CoTM~\cite{Thesis_CoalescedTM} applies a single clause pool, common for all $m$ outputs/classes. Different sets of weights, $w_{i,j}$, are applied to the clause outputs, $\{c_0, \dots, c_{n-1}\}$, to generate the class sums:
\begin{equation} \label{PaperD_Equation: Class Sum}
    v_i = \sum_{j=0}^{n-1} w_{i,j} c_j,
\end{equation}
where 
$i \in \{0, \dots, m-1 \}$ 
is the class index. 

The clause weights
in Eq.~(\ref{PaperD_Equation: Class Sum}) can have both positive and negative polarity, and obtain their values during training. \textit{No multiplications} are needed to implement Eq.~(\ref{PaperD_Equation: Class Sum}), 
as a clause{\color{black}'s output is} 
either~0 or~1. The predicted class, $\hat{y}$, is determined by the \textit{argmax} operator{\color{black}:}

\begin{equation} \label{PaperD_Equation: Class Decision by ArgMax}
    \hat{y} = \text{argmax}_{i}\{v_i\}, ~i\in\{0,\ldots, m-1\}. 
\end{equation} 

\subsection{CoTM and Convolution} \label{PaperD_subsection: CoTM and Convolution}

It has been shown that applying convolution to the TM can improve the image classification test accuracy~\cite{Thesis_CTM}. Convolution also reduces the number of features that need to be processed simultaneously, which in turn reduces model size and design complexity. However, the benefits of convolution come at the cost of increased processing time, due to the need to sequentially process the various patches of the input sample.

For image classification, the input data samples to a ConvCoTM, are a set of images, each of dimensions $X \times Y$, with $Z$ channels. The different patches of the image are generated by applying a sliding window of size $W_X \times W_Y$, with $Z$ channels. The convolution window will be evaluated \mbox{$B=B_X \times B_Y$} times across the image, where \mbox{$B_X=1+ (X-W_X)/d_x$}, \mbox{$B_Y=1+(Y-W_Y)/d_y$}, and $d_x$ and $d_y$ are the stride values of the convolution window in the $X$ and $Y$ directions respectively~\cite{Thesis_CTM}. For each window position, the number of Boolean features generated per patch
is given by 
\begin{equation} \label{PaperD_Equation: Number of features per patch}
    N_{F} = W_{X} \times W_{Y} \times Z \times U + (Y-W_{Y})+(X-W_{X}).
\end{equation}

In Eq.~(\ref{PaperD_Equation: Number of features per patch}), {\color{black}$N_F$ corresponds to $o$, which is the number of Boolean variables which are input to the TM, as described in Subsection~\ref{subsection General TM Concepts}.} $U$ is the number of bits used for thermometer encoding~\cite{Thesis_buckman2018thermometer} of the value of a single pixel (one-hot encoding can also be used). Such encoding is required because a TM operates on Boolean variables, not on binary numbers. The term $(Y-W_Y)+(X-W_Y)$ in Eq.~(\ref{PaperD_Equation: Number of features per patch}) represents the number of bits that encode the patch's position~\cite{Thesis_CoalescedTM, Thesis_10812055_TCSI_ConvCoTM28x28_FPGA} in the $Y$ and $X$ directions respectively. Also here we apply thermometer encoding.
As shown in Eq.~(\ref{PaperD_Equation: Literals}), the ConvCoTM also takes as input the negated versions of the features to form the \textit{literals} per patch.

\begin{figure*}[t] 
    \centering        \includegraphics[width=0.75 \textwidth]{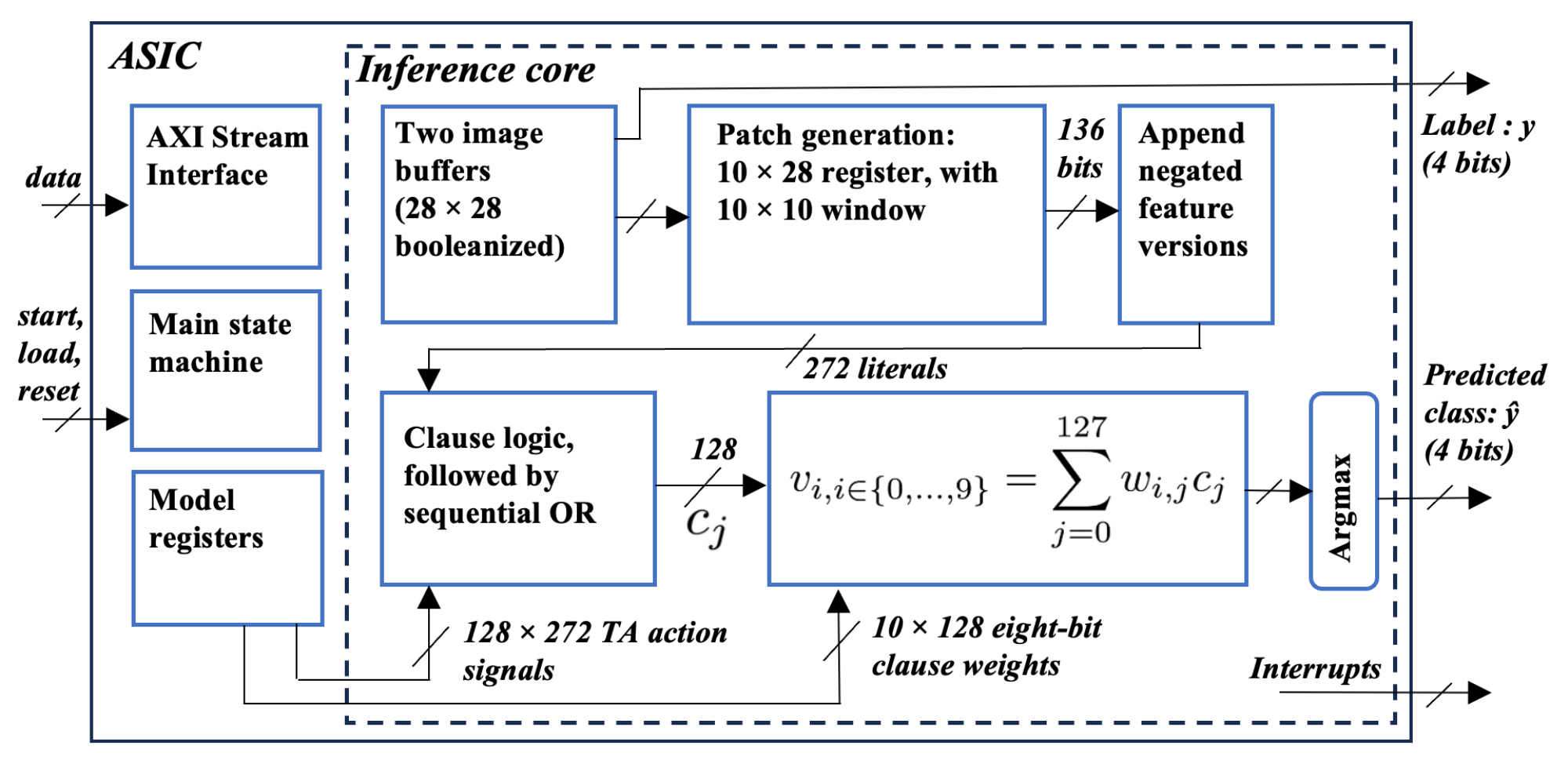}
    \caption{ConvCoTM accelerator ASIC block diagram.}  
    \label{PaperD_Figure: Accelerator block diagram}
\end{figure*}

\subsection{Datasets and Booleanization} \label{PaperD_subsection: Datasets and Booleanization}

The widely used MNIST 
dataset consists of handwritten digits. FMNIST 
is more challenging than MNIST and includes images from a Zalando catalog, such as t-shirts, sandals, and shoes. KMNIST 
consists of cursive Japanese characters, and is also more demanding than MNIST. Each of these datasets includes images of~$28 \times 28$ pixels organized in~60k training samples and~10k test samples. There are~10 classes within each dataset. As the images are of single-channel (greyscale) type,~$Z=1$.

The pixel values range from~0 to~255. For MNIST, they are converted into Boolean variables (i.e., \textit{booleanized}) with simple thresholding, where pixel values larger than~75 are replaced with~1, and~0 otherwise~\cite{Thesis_CTM}. For FMNIST and KMNIST, 
an adaptive Gaussian thresholding procedure~\cite{Thesis_CTM} is applied. We use one bit per pixel, i.e., \mbox{$U=1$} for all three datasets. The convolution window has dimensions $W_X=W_Y=10$, and a step size of one is used in both directions. For each patch, we {\color{black}therefore} have 100 feature bits from the convolution window plus~36 bits for the appended patch position information, ref. Eq.~(\ref{PaperD_Equation: Number of features per patch}). {\color{black}Thus, there are 272 literals per patch.}

\subsection{TM and Convolution - Inference}\label{PaperD_subsection: Convolution and Inference} 

During inference, each clause in a ConvCoTM 
is evaluated~$B$ times per image. After finishing the patch generation, a clause will output~1 if it has recognized a pattern \textit{at least once} in any of the $B$ patches for a given image~\cite{Thesis_CTM}. The clause output is given by Eq.~(\ref{PaperD_Equation CTMclauseOR}) which we {\color{black}denote as} 
the \textit{sequential OR-function}. Here~$c_j$ is the~$j$'th clause of the ConvCoTM, and~$c_j^b$ is the $b$'th output of this clause obtained during the patch generation:
\begin{equation} \label{PaperD_Equation CTMclauseOR}
    c_j = \bigvee_{b=0}^{B-1}c_j^b.
\end{equation} 

Class sum generation and class prediction are similar to the case without convolution, and follow Eq.~(\ref{PaperD_Equation: Class Sum}) and Eq.~(\ref{PaperD_Equation: Class Decision by ArgMax}).

\section{Design Overview} \label{PaperD_Section: Design Overview}

The block diagram of the ConvCoTM accelerator is shown in Fig.~\ref{PaperD_Figure: Accelerator block diagram}. The accelerator includes all circuitry required for performing inference, and only minimal interaction with the system processor is required. The three main modules are i)~a data
interface for transfer of model parameters and image samples, ii)~a module for model storage and iii)~an inference core that performs classification of an image. The following subsections describe the different sub-modules and the accelerator's  operation in detail.

\subsection{Interface to the System Processor}

An external system processor is required for i)~applying a few single-bit control signals to the accelerator, ii)~feeding data to the accelerator and iii)~reading interrupt signals and class predictions from the accelerator. All other operations are performed by the accelerator.

The 8-bit parallel data interface towards the external system processor, is inspired by the ARM AXI Stream interface~\cite{Thesis_AXI4StreamProtocol}. It is used for transferring model data to the accelerator during \textit{load model mode}, and for transferring images in \textit{inference mode}. When a classification operation is finished, the accelerator outputs an interrupt{\color{black}, and a}
separate 8-bit signal consisting of the predicted class
of the sample together with the true label (if provided).

\subsection{ConvCoTM Model Registers}

The model of a given CoTM configuration 
consists of:~i)~the TA action signals (\textit{include/exclude} signals) per clause and~ii)~the signed weights per clause per class. The TA action signals are stored in registers, readily available for simultaneous clause evaluation. These signals require~$272 \times 128=34816$ D flip-flops (DFFs). Registers are also applied for the clause weights per class. For this ASIC, eight bits are allocated per weight, with two's complement representation. The number of DFFs for storing the weights is $10 \times 128 \times 8=10240$, and the complete model size used by the accelerator is 45056 bits, i.e.,~5632 bytes. 

\subsection{Image Data Buffer and Patch Generation}\label{PaperD_subsec: data buffer and patch generation}

The accelerator includes a data buffer with room for two complete $28 \times 28 $ booleanized images. While one image is being processed, another image can be transferred from the external system processor to the other image register, thereby improving throughput. We denote this as \textit{continuous} classification mode. The size of a complete booleanized image is~$28 \times 28/8$ bytes $=98$ bytes. One additional byte is allocated for the image label~(0 to 9). 

\begin{figure}[!ht] 
    \centering
    \includegraphics[width=0.46 \textwidth]{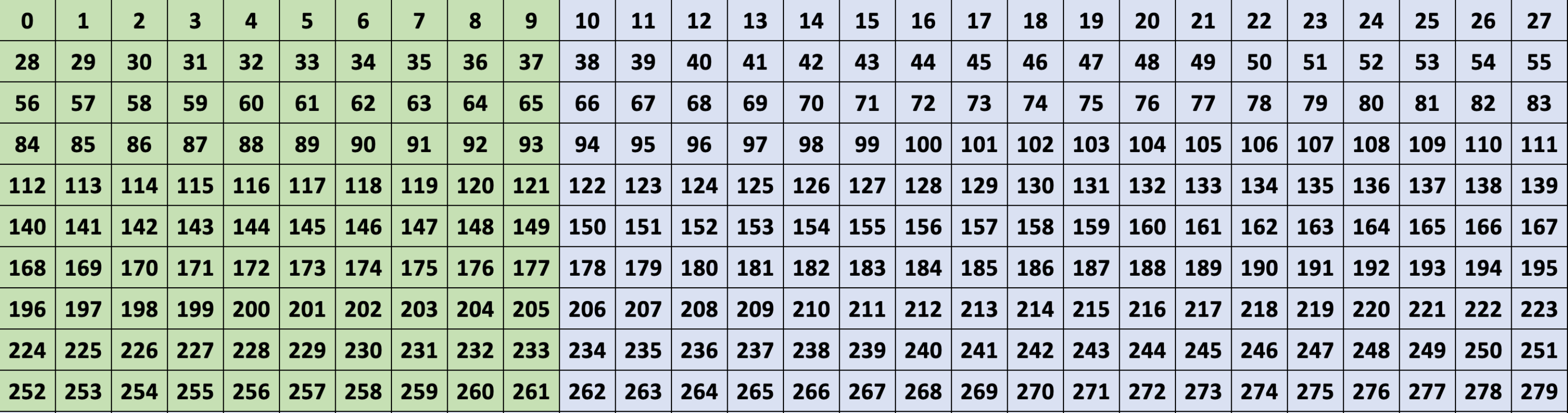}
    \caption{{\color{black}Register structure for window sliding. Each square represents a DFF. The first 10 rows of a booleanized image have been loaded and the window is in the start position (0,0).}}
    
    \label{PaperD_Figure: Patch generation 0}
\end{figure}

The patch generation module consists of a register with~10 rows, each with~28
DFFs. At the start of processing a new sample, the first~10 rows of an image are loaded into the register, and the convolution window starts at the leftmost position, which has coordinates ($0,0$){\color{black}, see Fig~\ref{PaperD_Figure: Patch generation 0}.}

During the patch generation, the window is moved to the right with increments of one. As described in Subsection~\ref{PaperD_subsection: CoTM and Convolution}, the position of the window is thermometer encoded. There are~19 different $x$-coordinates for the window, requiring~18 bits. When the window has been evaluated at the rightmost $x$-position~($0,18$), the contents of all~10 rows are shifted upwards with one step, and the next image datarow is loaded into the lowermost register row. The convolution window then starts again at the leftmost position, now with patch coordinates~($1,0$). This procedure is repeated until all~28 datarows of the image have been processed. There are~19 different positions also for the $y$-coordinate, and during a complete patch generation phase, $19 \times 19 =361$ patches are generated. {\color{black}The $x$- and $y$-positions are thermometer encoded, each with 18 bits as shown in Table~\ref{PaperD_Table: Position encoding}.} 

\begin{table}[!h]
\begin{center}
\caption{{\color{black}Thermometer position encoding of $10 \times 10$ convolution window in a $28 \times 28$~image. $x$ at position 0 means leftmost column, while $y$ at position 0 means uppermost row.}}\label{PaperD_Table: Position encoding}
\centering
\begin{tabular}{c|c}
\hline
$x$ or $y$ position & Thermometer encoded value (18 bits)\\
\hline
0 &  000000000000000000\\
\hline
1 & 000000000000000001\\
\hline
$\cdots$ & $\cdots$\\
\hline
17 & 011111111111111111\\
\hline
18 & 111111111111111111\\
\hline
\end{tabular}
\end{center}
\end{table}

\subsection{Clause Pool} \label{PaperD_subsec: Clause Pool}

\begin{figure}[!ht] 
    \centering
    \includegraphics[width=0.48 \textwidth]{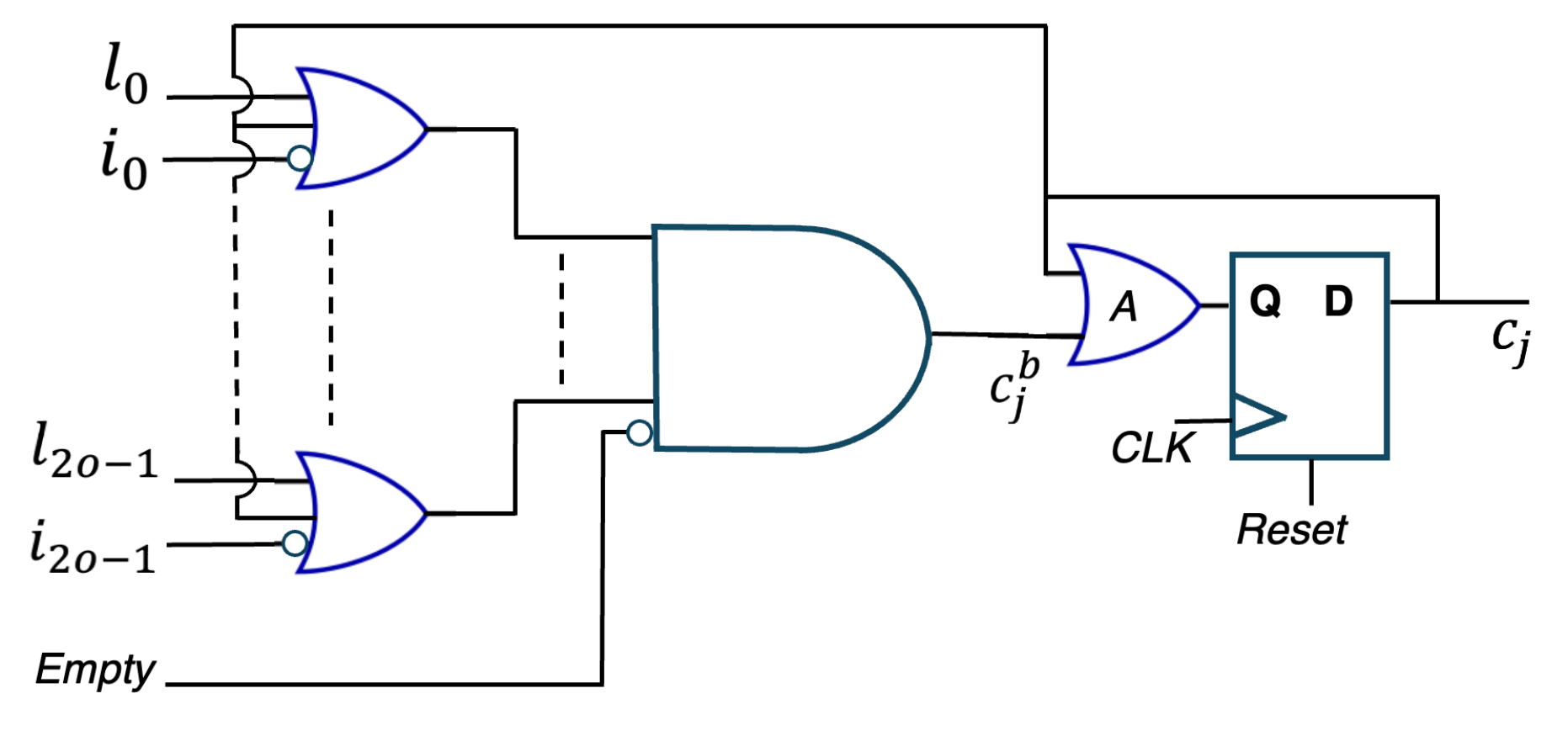}
    \caption{Circuitry for a single clause. Literals are denoted with $l$ and TA-actions with $i$. 
    }
    \label{PaperD_Figure: Clause_logic_with_seq_OR}
\end{figure}

The ConvCoTM is configured with 128 clauses that operate in parallel. Each clause takes the
literals, $[l_0, l_1, \dots, l_{271}]$, as input, as well as their corresponding \textit{TA action} signals, $[i_{j,0}, i_{j,1}, \dots, i_{j,271}]$, $j \in \{0,\ldots, 127 \}$, from the  
model register. Fig.~\ref{PaperD_Figure: Clause_logic_with_seq_OR} shows a simplified circuit diagram of the clause logic. A literal
will be included in the logical AND expression, that constitutes the \textit{clause}, if its corresponding TA action signal is high. A special situation occurs when a clause is \textit{empty}, i.e., when there are no \textit{include} TA actions. In this case, additional logic will set the signal \textit{Empty} high, forcing~$c_j^b$ low. 

The DFF in Fig.~\ref{PaperD_Figure: Clause_logic_with_seq_OR} constitutes a single-bit register for \mbox{clause $j$}. Its content, $c_j$, is ORed with the combinational clause value, $c_j^b$, of the current patch, during the patch generation. This implements the \textit{sequential OR function} in Eq.~(\ref{PaperD_Equation CTMclauseOR}). The DFF is reset before starting a new convolution process. 

The signal $c_j$ is also fed back to the input of the OR-gates, as there is no need for the combinational logic to evaluate a clause for more patches if $c_j$  already is high. This feedback, hereinafter denoted the \textit{clause switching reduction feedback}~(CSRF), reduces the switching in the combinational part of the clause logic. Simulations of MNIST classification utilizing CSRF, showed an average of~50\% reduction in the toggling rate of~$c_j^b$ for each clause. The accelerator ASIC has a dedicated pin that can enable or disable the CSRF. 

In this accelerator ASIC we apply a single convolution window. If increased classification rate is wanted, one can use several convolution windows that operate in parallel. In this case, the combinational clause logic that generates the signal~$c_j^b$ in~Fig.~\ref{PaperD_Figure: Clause_logic_with_seq_OR}, would have to be replicated for each convolution window. {\color{black}The outputs from each combinational clause logic instance would have to be input to the OR-gate labeled A in~Fig.~\ref{PaperD_Figure: Clause_logic_with_seq_OR}.} 

\subsection{Inference}\label{PaperD_subsection: Design Overview Inference}


The inference operation starts with loading an image into the data buffer, followed by sequential generation of the patches, as described in Subsection~\ref{PaperD_subsec: data buffer and patch generation}. When finished, each clause value, $c_j$, is weighted, for each class, according to Eq.~(\ref{PaperD_Equation: Class Sum}).
All ten class sums are generated in parallel{\color{black}, and each class sum is obtained by adding 128 numbers as shown in Fig.~\ref{PaperD_Figure: Class sum generation}. {\color{black}For each adder input, a multiplexer (MUX) is employed, which outputs 
the clause weight, $w_{i,j}$, if $c_j=1$, or zero if $c_j=0$. 
The adders are} configured in a {\color{black}reduction}-tree, with a three-stage pipeline. 

\begin{figure}[!ht] 
    \centering
    \includegraphics[width=0.45\textwidth]{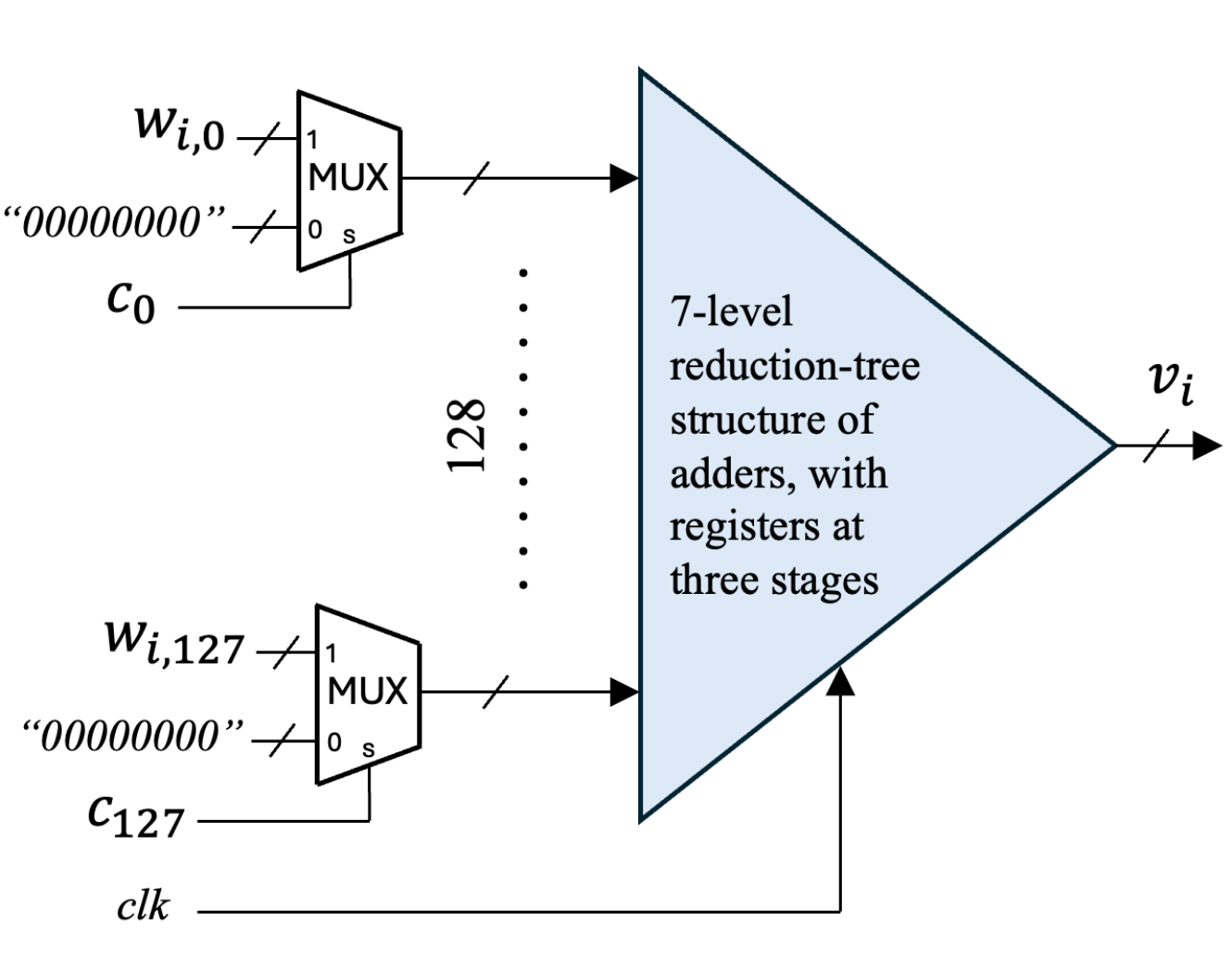}
    \caption{{\color{black}Class sum generation for class $i$. The main input signals are the class weights, $w_{i,j}$, and the Boolean clause values, $c_j$, where $j\in\{0, \dots, 127\}$.}}
    \label{PaperD_Figure: Class sum generation}
\end{figure}

{\color{black}Finally, an \textit{argmax} module selects the class with the highest class sum, as the \textit{predicted class} $\hat{y}$. Fig.~\ref{PaperD_Figure_argmax_schematic} shows the \textit{argmax} module which is implemented by combinational logic, using a submodule from~\cite{Thesis_Wheeldon1} in a reduction-tree structure. In Fig.~\ref{PaperD_Figure_argmax_schematic}, this submodule is shown in the upper right part. 
The inputs are two class sums, $v_0$ and $v_1$, 
and their corresponding \mbox{4-bit} class labels, \textit{label0} and \textit{label1}. If $v_1 > v_0$, the module feeds forward~$v_1$ as the $v_{max}$ result, and \textit{label1} is output as the maximum argument,~$a_{max}$. Otherwise, $v_0$ and \textit{label0} are selected. 
}



\begin{figure}[!ht] 
    \centering
    \includegraphics[width=0.50\textwidth]{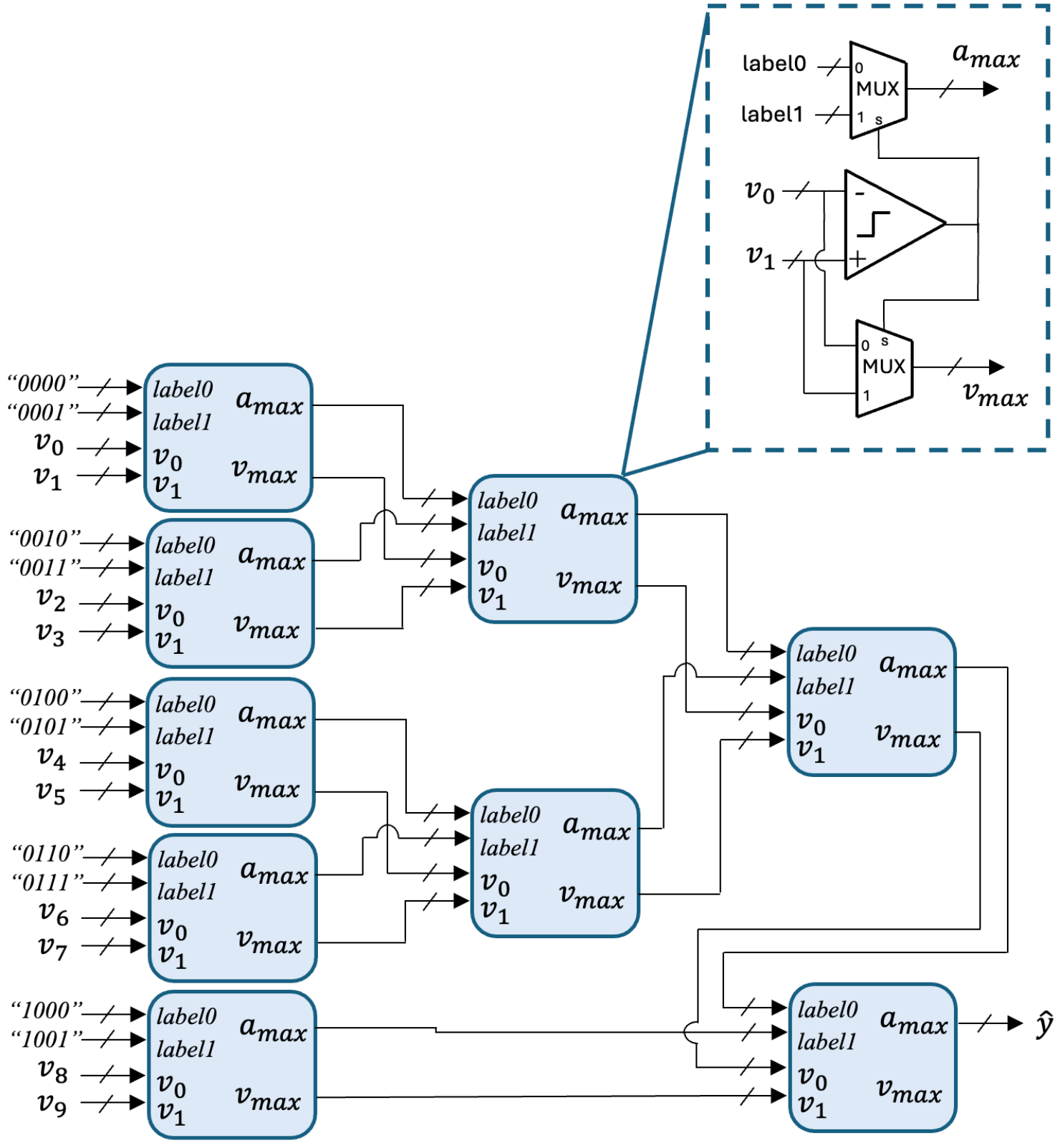}
    \caption{{\color{black}The \textit{argmax} module, that selects the label corresponding to the largest class sum. Based on a submodule circuit from~\cite{Thesis_Wheeldon1}.}}
    \label{PaperD_Figure_argmax_schematic}
\end{figure}

Fig.~\ref{PaperD_Figure: Inference State diagram} shows a simplified state diagram of the accelerator, and pseudo-code for the inference procedure is described in Algorithm~\ref{PaperD_alg:inference}. 

\begin{algorithm*}[!ht]
 \caption{{\color{black}ConvCoTM Accelerator Inference}}\label{PaperD_alg:inference}
  \begin{algorithmic}[1]
  \scriptsize
    \Procedure{Classify}{$A$} \Comment{$A$ is a single image example fed to the accelerator by the system processor.}
    \State \Comment{From here the algorithm is performed in its entirety by the accelerator.}
    \State \texttt{Reset \textit{clause output register}.}\Comment{Resets the DFF in each instance of the clause logic, ref. Fig.~\ref{PaperD_Figure: Clause_logic_with_seq_OR}.} 
      \For{\texttt{$i = 0$ to $B-1$}} \Comment{Process $B$ patches, ref. Subsections  \ref{PaperD_subsection: CoTM and Convolution} and \ref{PaperD_subsec: data buffer and patch generation}.}
        \State \texttt{\textbf{Perform in parallel during a single clock cycle:}}
        \State \texttt{* Generate patch($i$) of $A$}
        \State \texttt{* Evaluate all clauses, $c_j$, $j \in \{0, \dots, n-1\}$, for patch($i$)}
        \State \texttt{* OR the \textit{clause output register $c_j$ with the combinational clause output $c_j^b$}} 
        \State \texttt{\hspace{9pt}and update \textit{clause output register}} \Comment{Ref. Eq. (\ref{PaperD_Equation CTMclauseOR}) and Fig.~\ref{PaperD_Figure: Clause_logic_with_seq_OR}.}
      \EndFor
      \State \texttt{Calculate class sums: $v_0, v_1, \dots, v_{m-1}$} \Comment{Ref. Eq.~(\ref{PaperD_Equation: Class Sum}). A tree structure of adders, pipelined in~3 stages, is used per class sum.}
      \State \textbf{return} \texttt{argmax\{$v_0, v_1, \dots,  v_{m-1}$\}} \Comment{Returns predicted class. Ref. Eq.~(\ref{PaperD_Equation: Class Decision by ArgMax}).}
    \EndProcedure
  \end{algorithmic}
\end{algorithm*}

\begin{figure}[!ht] 
    \centering
    \includegraphics[width=0.50\textwidth]{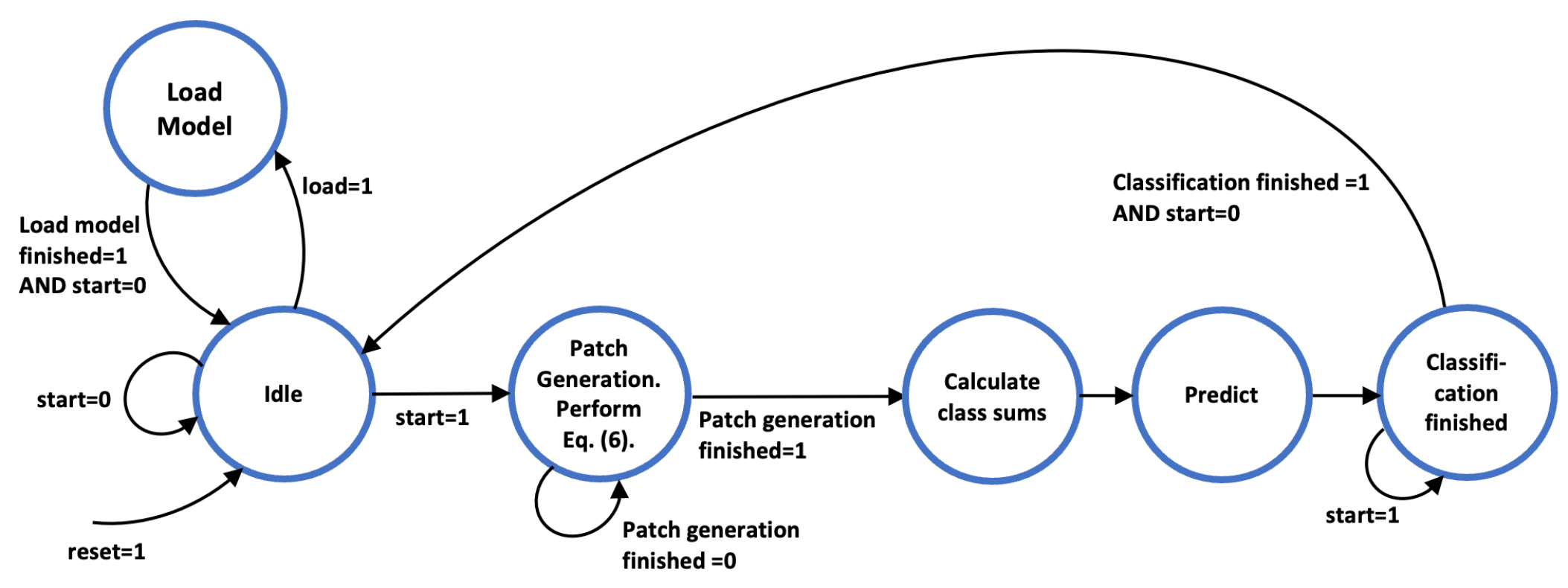}
    \caption{Simplified accelerator state diagram.}
    \label{PaperD_Figure: Inference State diagram}
\end{figure}

The accelerator's latency is 471 clock cycles. This is measured from when the system processor initiates the first byte transfer of the image, until a prediction is available.  It includes~99 clock cycles for transferring the~98 image bytes and the label byte, and~372 clock cycles for the patch generation, class summation and class prediction.

When the accelerator operates in continuous mode, a new sample is loaded into the image buffer while the current sample is being processed, see Subsection~\ref{PaperD_subsec: data buffer and patch generation} {\color{black}and Fig.~\ref{PaperD_Figure: Inference timing diagram}}. This increases the throughput, as image samples can be processed every 372'th clock cycle. Any timing overhead in the system processor will add to the total latency.

\begin{figure}[!ht] 
    \centering
    \includegraphics[width=0.4\textwidth]{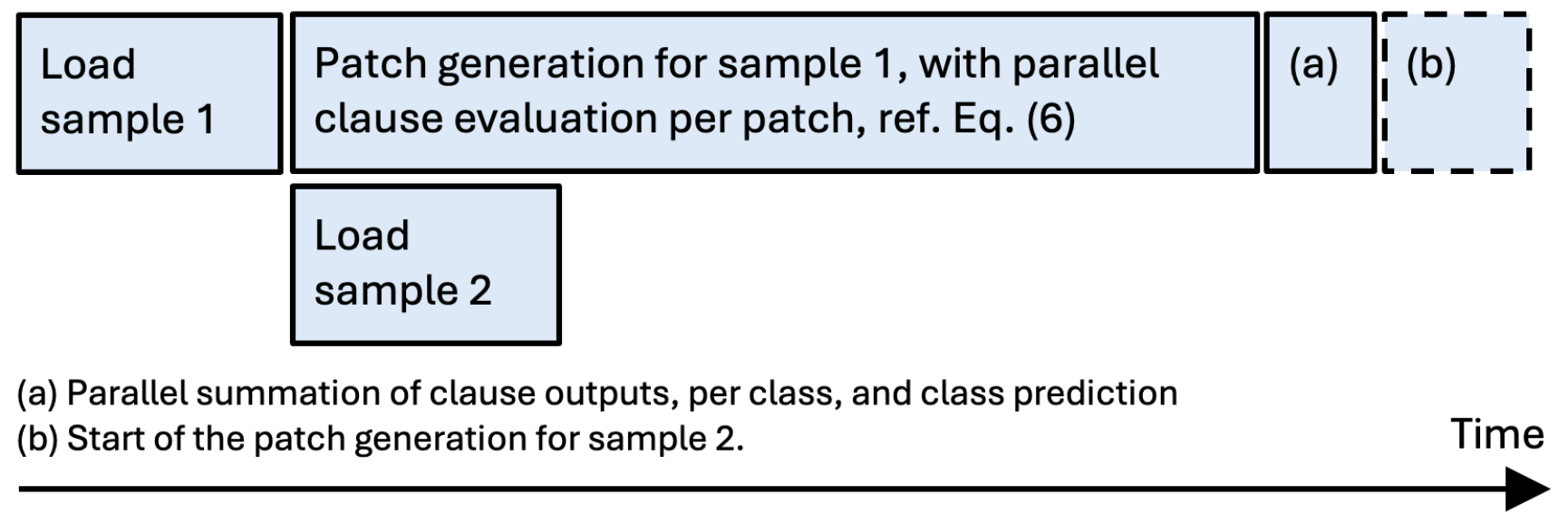}
    \caption{{\color{black}Simplified accelerator timing diagram.}}
    \label{PaperD_Figure: Inference timing diagram}
\end{figure}

\subsection{Clock Domains}

The key to reducing the digital switching power of the accelerator, lies at the top level of the ASIC architecture. Two separate clock domains are utilized, each with its own dedicated clock pin, assigned to the model part and the inference core, respectively.

When a model has been transferred from the system processor to the ASIC, the TA action signals and the weights are available in registers that do not need further clocking. By stopping the clock to the model part, when operating in inference mode, the digital switching power is reduced significantly, as the model part constitutes 
approximately 90\% of the accelerator's~DFFs. To further reduce power consumption, standard clock gating is applied for the inference module. This ensures, for instance, that the pipelined registers associated with the class sum generation are enabled and clocked only for four clock cycles per classification phase. The clock-gating can be enabled/disabled by an external pin. 
\section{Implementation and Measurement Results}\label{PaperD_Section: Implementation and Measurement Results}

\begin{figure*}[!ht] 
    \centering
    \includegraphics[width=0.8\textwidth]{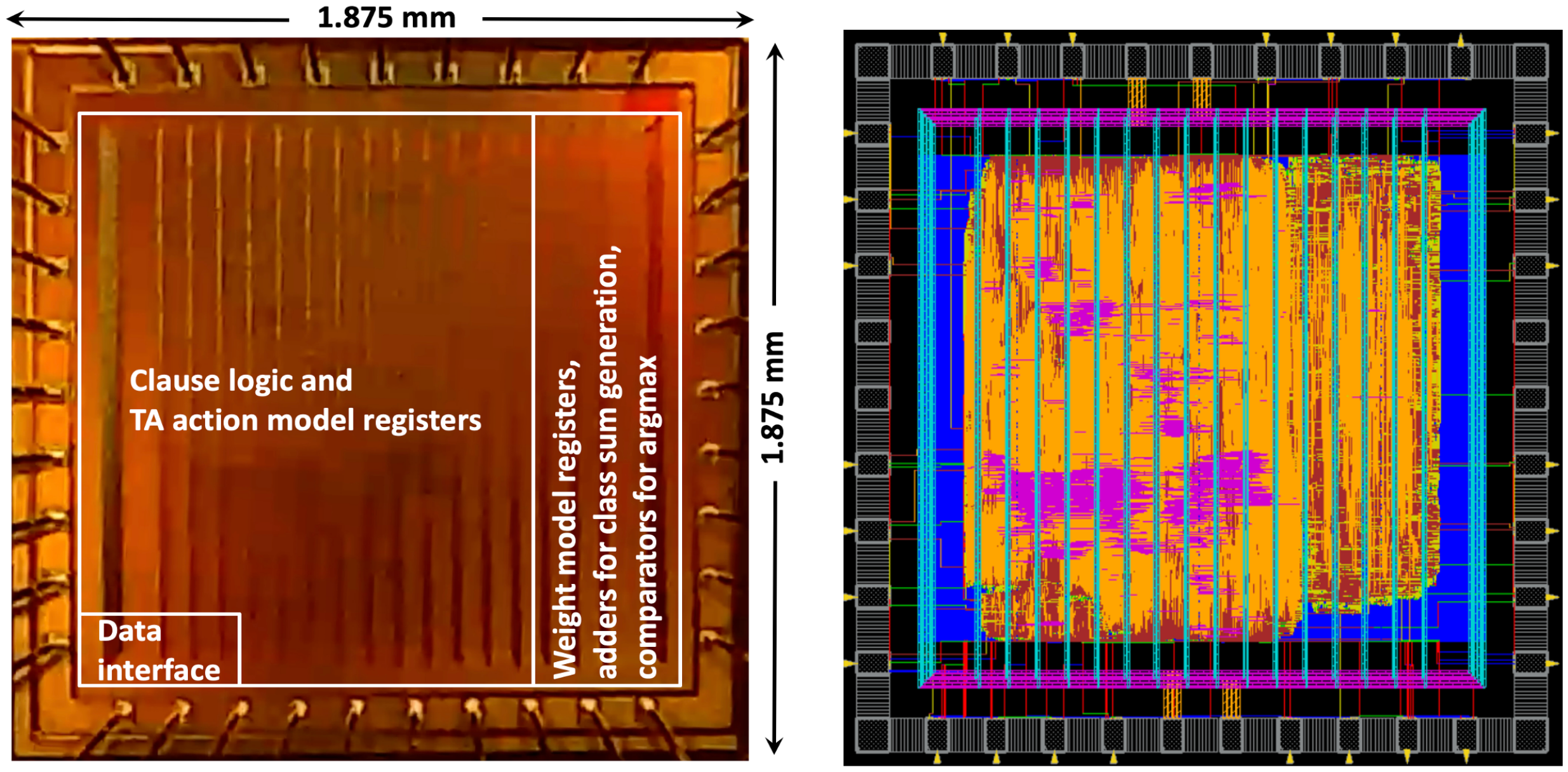}
    \caption{Photo and layout plot of the ConvCoTM accelerator chip.}
    \label{PaperD_Figure: ASIC microphotography}
\end{figure*}

The ConvCoTM accelerator was designed in VHDL and implemented in a~65~nm low-leakage CMOS technology from UMC. {\color{black}We used the \textit{Cadence} IC design framework for the chip implementation, and the VHDL simulations were performed with \textit{Xcelium}. Synthesis and Place~\&~Route were performed with \textit{Genus} and \textit{Innovus} respectively. To reduce leakage current and to optimize the power consumption for low operating frequencies, relaxed time-constraints were applied, and digital cells with high drive strengths were excluded. 

For voltage-drop analysis we applied \textit{Voltus}, and 
\textit{Conformal} was used for logic equivalence checking. For timing verification of the routed design, we used mixed VHDL/Verilog gate-level simulations with back-annotated timing information. Complete gate-level simulations of the accelerator with model loading and inference of the full MNIST test dataset were performed. Design-rule check (DRC) and layer versus schematic (LVS) were performed with \textit{Calibre}, while GDSII file generation was performed with \textit{Virtuoso}. The VHDL source code for the ASIC implementation (used for initial FPGA verification) is publicly available at a GitHub repository~\cite{Thesis_GitHub_ASIC_repository}.}

A~JLCC44 ceramic package housed the chip. {\color{black}For the package pins, there were~5 ground connections,~2 supply voltage connections (3.3~V) for the digital inputs/outputs (IOs),~2~supply connections (1.2~V) for the core cells, and~29~IOs. Six pins were not connected. The number of pins could have been reduced significantly, but as the ASIC was a test chip, we prioritized testability and observability.} Fig.~\ref{PaperD_Figure: ASIC microphotography} shows a photograph and a layout plot of the ASIC. The accelerator's core area is approximately~2.7~mm$^2$. {\color{black}For the testing and characterization of the ASIC, a printed circuit board (PCB) was developed.} The ASIC PCB was connected to a Digilent \mbox{Zybo Z7-20} development board,  equipped with a Xilinx Zynq XC7Z020 FPGA, as shown in Fig.~\ref{PaperD_Figure: Test setup}. One of the FPGA's ARM9 cores was configured as system processor, and the FPGA board provided clocks, model data, image data samples and control signals to the ASIC PCB. The processor also read back accelerator status signals, and the predicted class per sample. To interface with the FPGA board, the ASIC's IOs  operated at~3.3~V supply voltage. The accelerator core operated from~0.82~V to~1.2~V. 

\begin{figure}[!ht] 
    \centering
    \includegraphics[width=0.43\textwidth]{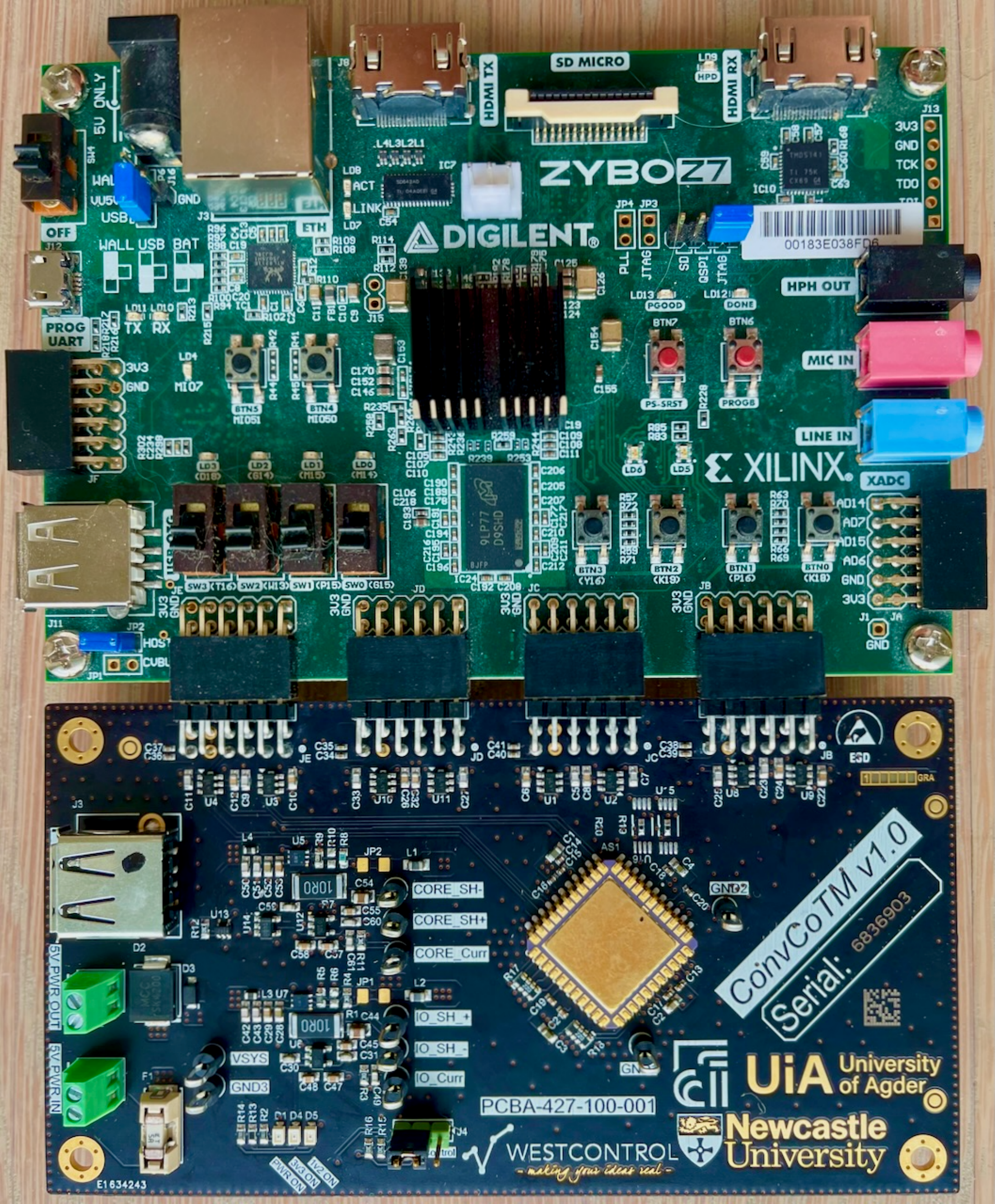}
    \caption{Test setup with the ConvCoTM ASIC test board connected to the FPGA development board.}
    \label{PaperD_Figure: Test setup}
\end{figure}

For each of the datasets MNIST, \mbox{FMNIST} and KMNIST, the \textit{Tsetlin Machine Unified}~(TMU)~SW-version~\cite{Thesis_TMU_CoalescedClassifier} of the ConvCoTM was trained to find suitable models. Maximum/minimum limits were set on the clause weights to fit with the allocated 8-bits per weight in the accelerator. For simplicity, the images from the test datasets were preprocessed (booleanized), as described in Subsection~\ref{PaperD_subsection: Datasets and Booleanization}, and included in the FPGA processor's test program. 

Model data and image samples were applied from the FPGA to the accelerator via the AXI Stream interface. We utilized direct memory access (DMA) functionality on the FPGA to enable fast data transfer. The maximum classification rate achieved, including system processor timing overhead, was~60.3~k images per second in continuous mode, with an accelerator clock frequency of~27.8MHz. The maximum operating frequency was limited by the general purpose IOs (GPIOs) on the FPGA board. When classifying a single image, the latency was~25.4~$\mu$s, including the transfer time of the image to the accelerator as well as timing overhead in the system processor. 

The achieved test accuracies of the datasets \mbox{MNIST}, \mbox{FMNIST} and \mbox{KMNIST} were 97.42\%,~84.54\% and~82.55\% respectively. These results are independent of the accelerator clock frequency and are exactly in accordance with the performance of the models obtained from the SW~simulations. 

\begin{table}
\begin{center}
\caption{The ConvCoTM accelerator ASIC's main characteristics and performance}\label{PaperD_Table: accelerator characteristics and performance}
\centering
\begin{tabular}{l|c}
\hline
\textbf{Parameter / Characteristics} & \textbf{Value / Description}\\
\hline
Technology & 65 nm low-leakage CMOS from UMC\\
\hline

Chip area (full die) & 3.5 mm$^2$\\
\hline

Chip area (core) & 2.7 mm$^2$\\
\hline

Gatecount (core) & 201k cells\\
                 & including 52k DFFs \\
\hline
Power consumption   & 1.15~mW $^{a,c}$\\
(accelerator core only)& 0.52~mW $^{a,d}$\\
  & 81~$\mu$W $^{b,c}$\\
  & 21~$\mu$W $^{b,d}$\\

\hline

Classification rate & 60.3~k images/s $^{a}$  \\
 (including system overhead) & 2.27~k images/s $^{b}$ \\

\hline

Energy per classification & 19.1 nJ $^{a,c}$\\
(accelerator core only) & 8.6 nJ $^{a,d}$\\
 & 35.3 nJ $^{b,c}$\\
 & 9.6 nJ $^{b,d}$\\
\hline

Latency (single image classi-  & 25.4 $\mu$s $^{a}$\\
fication including data transfer) & 0.66 ms $^{b}$\\

\hline
Test accuracy & 97.42\% (MNIST)\\
& 84.54\% (FMNIST)\\
& 82.55\% (KMNIST)\\

\hline
\end{tabular}
\end{center} 
$^a$~27.8~MHz clock frequency. $^b$~1.0~MHz clock frequency. $^c$~The accelerator core supply voltage (vdd) is~1.20~V. $^d$~vdd=0.82~V.
\end{table}

For power consumption measurements, a Joulescope JS220 precision energy analyzer 
was used. 
The power consumption during inference was measured while the accelerator was set in a test mode with repeated classifications of the full~10k sample dataset. The power consumed by the accelerator core with~1.20~V supply voltage and a clock frequency of~27.8~MHz was~1.15~mW. This corresponds to an EPC of~19.1~nJ. 
With a supply voltage of~0.82~V, the accelerator core's power consumption was~0.52~mW and the corresponding EPC was~8.6~nJ. 

Table~\ref{PaperD_Table: accelerator characteristics and performance} summarizes the characteristics and performance of the accelerator. In a real application, the ConvCoTM design would typically be included as a peripheral module to a microcontroller in an SoC{\color{black}, and the digital IO interface would have been optimized}. Therefore, we considered the power consumption of the accelerator \textit{core} as the most interesting power parameter. In inference mode, the digital~3.3~V~IOs, which were required for interfacing to the FPGA board, had a power consumption of~0.76~mW, at a clock frequency of~27.8~MHz. 

All {\color{black}reported} power and energy measurement results of the ASIC are average values, and were performed with clock-gating and CSRF enabled (as shown in Fig.~\ref{PaperD_Figure: Clause_logic_with_seq_OR}). When operating at~27.8~MHz, clock-gating reduced the power consumption by approximately~60\%, while the CSRF alone provided less than~1\% power reduction.
{\color{black}
\section{Possible Extensions}\label{PaperD_Section: Possible extensions}
In this section, we describe a technique which limits the number of literals that are input to each clause~\cite{Thesis_ijcai2023p378_constraining_clause_size}. This can enable a significant reduction in model size and chip area, and we provide estimates for a~28~nm CMOS implementation of the ConvCoTM accelerator. We also discuss how the accelerator could be modified to support on-device training. Finally, we describe concepts and estimates for a scaled-up solution intended for operation on the CIFAR-10 dataset.}

{\color{black}
\subsection{Estimated Accelerator Performance if Implemented in~28~nm CMOS}\label{PaperD_subsec_ASIC_scaled_to_28nm}

In the current ConvCoTM ASIC accelerator, each clause takes all the literals as input, ref. Fig.~\ref{PaperD_Figure: Clause_logic_with_seq_OR}. This makes it necessary to store and use also the \textit{exclude} TA actions of the model. However, for a given clause, the literals that are excluded do not contribute to the the clause value generation and therefore represent an overhead of the model. 





TM models are in general highly sparse. For example, in the MNIST model used for the ConvCoTM accelerator,~88\% of the TA actions are \textit{exclude}. There is active research on how to utilize the TM model sparsity to achieve more compact HW and SW solutions. One option is to use a TM training setting that specifies a maximum number of literals per clause~\cite{Thesis_ijcai2023p378_constraining_clause_size}, with only negligible loss of accuracy. This enables a significant reduction of the model size and the associated chip area for storage and processing of the TA actions. 

As an example, we here consider the current ASIC and assume that each clause only takes~10 literals as input. To select a single literal, one would need a \mbox{272-to-1} multiplexer, and a \mbox{9-bit} \textit{literal address} (multiplexer selection word). That is, one would need $10 \times 9=90$ bits per clause, for storing which literals are included. The principle for the clause logic could be as shown in Fig.~\ref{PaperD_Figure_Clause_logic_with_reduced_no_literals}. In this case, an area reduction of~$(272-90)/272 \approx 67$\% of the TA action part of the model is achieved. In the current ASIC, the TA action part of the model, and the associated clause logic, constitute about~70\% of the accelerator core's chip area. Using the principle in Fig.~\ref{PaperD_Figure_Clause_logic_with_reduced_no_literals}, one could achieve a total reduction of the core area of approximately~47\%.


\begin{figure}[!ht] 
    \centering
    \includegraphics[width=0.5 \textwidth]{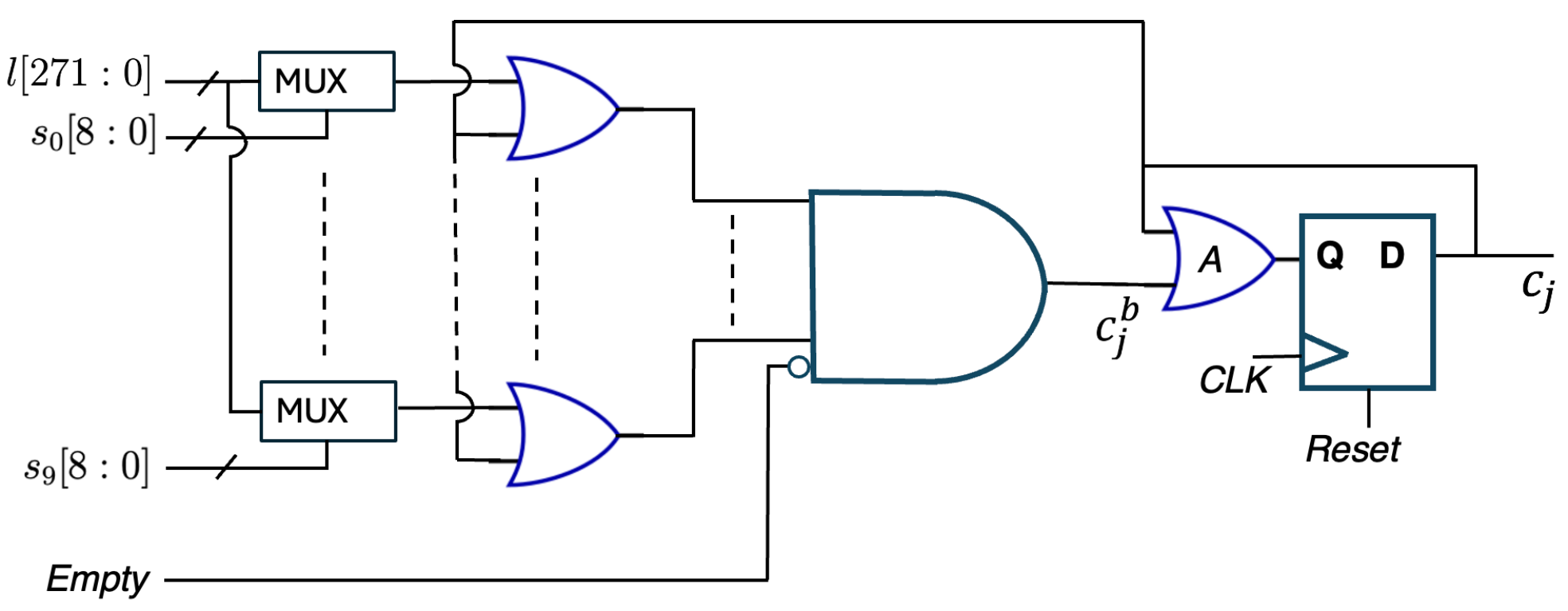}
    \caption{{\color{black}Circuitry for a single clause, where only 10 literals are used as input. $s_0$ to $s_9$ are multiplexer selection signals given by the model. 
    }
    }
    \label{PaperD_Figure_Clause_logic_with_reduced_no_literals}
\end{figure}

A reduction in chip area and EPC can be achieved by implementing the ConvCoTM accelerator in a more advanced technology. We here estimate the chip area, power consumption and EPC if a 28~nm CMOS technology was used. With Dennard scaling~\cite{Thesis_782564_Dennard_scaling_Borkar}, we roughly estimate a general reduction factor of the active chip area of $(28/65)^2$. In addition, we assume the number of literals per clause is limited to~10 as described above. Thus gives an estimated chip area of~0.27~mm$^2$. At an operating frequency of~27.8~MHz (used here for comparison as this was the maximum clock rate tested for the manufactured ConvCoTM ASIC), and~0.7~V supply voltage, we roughly estimate a~50\% reduction in power consumption compared to the~65~nm chip operating at~0.82~V. This results in an EPC of~4.3~nJ which would be close to what is achieved for the SNN accelerator in~\cite{Thesis_10902457_3p32nJ_per_Frame}. 
}

{
\color{black}
\subsection{Extension of the ASIC Design to Support Training}\label{PaperD_subsec_ASIC training}
}

{\color{black} 
Based on the ConvCoTM training concept and circuit modules used in the FPGA accelerator in~\cite{Thesis_10812055_TCSI_ConvCoTM28x28_FPGA}, we here describe how the ASIC could be modified to support on-device training.

During the patch generation phase, one has to perform random patch selection per clause. In~\cite{Thesis_10812055_TCSI_ConvCoTM28x28_FPGA}, this is performed based on the reservoir sampling algorithm~\cite{Thesis_knuth97}. One would need one or more RAM modules for storing all of the 361 generated patches, each consisting of 136 feature bits. In addition, a register of 9 bits is required per clause to store the address of the randomly selected patch. 

For each clause, 272 TAs would be needed, i.e., one TA for each literal. We here assume that 8-bit TAs are used, based on the structure shown in Fig.~\ref{PaperD_Figure_TA}. 
In~\cite{Thesis_10812055_TCSI_ConvCoTM28x28_FPGA}, all TAs belonging to a single clause are updated simultaneously. Therefore, the TAs can be implemented by parallel operation of~34 single-port RAM modules, each with a word width of 64 bits, supporting 8~TAs. The number of rows in the TA~RAMs would be 128, corresponding to the number of clauses in the ConvCoTM configuration used.


The model registers for the weights and the TA action signals, are already included in the ASIC, but to support training, register update logic would be required.

For random number generation during the reservoir sampling and during the clause updating phase, linear feedback shift-registers (LFSRs) of minimum 16 bits, are needed~\cite{Thesis_10812055_TCSI_ConvCoTM28x28_FPGA}. During the clause updating, one LFSR is used for the random decision of whether a clause's TAs should receive update, and 272 LFSRs are required for the simultaneous updating of all TAs per clause. 

The additional circuit modules required for implementing the training modules is moderate, and we estimate it to approximately~1~mm$^2$. For inference, the power consumption should not increase much, provided clock-gating is applied for the specific training modules, and RAM modules with very low leakage current are used for the TAs. Reading from and writing to the RAMs consume energy, but these operations would only be  performed during training. The accelerator in~\cite{Thesis_10812055_TCSI_ConvCoTM28x28_FPGA} operated at~50~MHz and achieved a training throughput of approximately~40k samples per second. For a modified ASIC that supports training, and operates at 27.8~MHz, we estimate a training throughput of~22.2k samples per second.\\
}

\subsection{Scaled-Up Solution}\label{PaperD_subsection_ScaledUp_CIFAR10}

The ConvCoTM accelerator's architecture described in this paper, is highly suited for scaling, and can be modified to operate on larger and more complex images. There is active research on how to apply the TM algorithm {\color{black}on} larger and more complex images. The TM version that currently achieves the best test accuracy on CIFAR-10 is 
\textit{TM Composites}~\cite{Thesis_granmo2023tmcomposites, Thesis_grønningsæter2024optimizedtoolboxadvancedimage}. Here, several \textit{TM Specialists} are applied to an image sample. Examples of \textit{specializations} are different booleanization techniques and convolution window sizes. After an image has been processed, the class sums from each TM Specialist are normalized, followed by summation per class. Finally, an \textit{argmax} operator is applied on the \textit{composite class sums} for output prediction.  

Table~\ref{PaperD_Table: CIFAR10 ASIC solution} shows characteristics and performance estimates for an envisaged TM Composites ASIC design, intended for inference operation on \mbox{CIFAR-10} images.
It is assumed that one configurable TM module is implemented in the ASIC, and models for four different TM Specialist configurations are applied in sequence, through transferring a new model from on-chip ultra-low-power RAM per configuration. 

Booleanization of the images, for the different TM Specialists, 
is assumed to be performed
by dedicated logic in the HW accelerator. 
We assume a timing-optimized data interface, with line buffers~\cite{Thesis_8746699_line_buffer}, allowing the patch generation to start as soon as a sufficient number of image lines have been transferred to the accelerator. 

For the TM Specialists, we assume an average number of literals per patch of~1000, and by utilizing the technique described in~\cite{Thesis_ijcai2023p378_constraining_clause_size}, the number of literals is limited to~16 per clause. For the TA actions, the model for each TM Specialist therefore needs to include~16 ten-bit literal addresses per clause. 

We first assume the same 65~nm CMOS technology is used as for the current ConvCoTM ASIC. The chip area for the \mbox{CIFAR-10} accelerator is estimated by multiplying the chip area of the current ASIC (2.7~mm$^2$) with the ratio, $R$, given by the model size of one TM Specialist (32.5~kilobyte{\color{black}s}) divided by the model size of the ASIC in this paper (5.6~kilobyte{\color{black}s}). We therefore get $R=32.5/5.6 \approx 5.8$. This is a reasonable assumption because the model storage in registers (for the active TM Specialist) and the clause logic, dominate the chip area. We further assume 2~mm$^2$ in additional chip area for booleanization logic, additional adders for class sum generation, and RAM modules for model storage.

We estimate that each TM Specialist will need approximately~1000 clock cycles for the processing per sample, including booleanization.
In addition, before a TM Specialist can start operating, its model has to be loaded from on-chip RAM. We assume the model is organized in memory such that a transfer rate of 32 bytes per clock cycle is enabled. Thus, we will need approximately~1020 clock cycles for a complete model transfer. A single TM Specialist will therefore require in total about 2020 clock cycles to process a sample, and with four TM Specialists we will need approximately~8080 clock cycles for a complete classification. With a system clock of~27.8~MHz (as for the current ConvCoTM ASIC), the classification rate would be \mbox{$27.8 \times 10^6/8080$~FPS~$\approx 3440$}~FPS.

An estimate of the power consumption is found by multiplying the power consumption of the current ASIC's accelerator core, operating at~27.8~MHz {\color{black}and~0.82~V}, with the factor $R$. 
During model loading and booleanization, we assume power consumption similar to that during inference. An estimate of the {\color{black}EPC} is then $R\times {\color{black}0.52}$~mW/(3440 FPS)~$\approx {\color{black}0.9}~\mu$J/frame.

{
\color{black}

For a 28~nm CMOS implementation of this solution, we roughly estimate an accelerator core area of~3.3~mm$^2$, based on Dennard scaling. Assuming 0.7~V supply voltage, a coarse estimate of the EPC is~0.45~$\mu$J. The operation frequency could be increased for higher throughput, and this could enable further reduction of the EPC.
}

\begin{table}[!h]
\begin{center}
\caption{Envisaged ConvCoTM inference ASIC Configuration and Performance for operation on the \mbox{CIFAR-10} Dataset, based on the TM Composites architecture.}\label{PaperD_Table: CIFAR10 ASIC solution}
\footnotesize
\centering
\begin{tabular}{l|c|l}
\hline
\textbf{Parameter} & \textbf{Value} & \textbf{Comment} \\
\hline
Number of TM & 4 & $4 \times 4$ color thermometer,\\
specialists &  & $3 \times 3$ color thermometer,\\
 &  &  $32 \times 32$ histogram of gradients,\\
 &  &  $10 \times 10$ adaptive thresholding.\\
\hline

Number of clauses & 1000 &  \\
\hline


Number of included & 16 & We assume 1000 literals\\
literals per clause &  & per patch, and literal limitation\\
 &  & based on~\cite{Thesis_ijcai2023p378_constraining_clause_size}.\\
\hline

Model size: TA actions & 20 & 1000 clauses, each with 16\\
per TM Specialist & kilobyte{\color{black}s} &literals. The address\\
& & length per literal is 10 bits.\\

\hline
Model size: Weights& 12.5 & 10 classes, 1000 clauses \\
per TM Specialist & kilobyte{\color{black}s} & and 10-bit weights.\\
\hline
Complete model size & 130 & For four TM Specialists.\\
& kilobyte{\color{black}s} & \\

\hline

Classification rate & 3440 FPS & Sequential operation of the\\

 &  & TM Specialists, 27.8 MHz \\
 &   & system clock, complete model \\
&  & stored on-chip, without \\
&  & system processor overhead. \\

\hline

Test accuracy & 79\% & Estimate for \mbox{CIFAR-10}.\\
\hline
Accelerator core & 17.7~mm$^2$ & 65~nm CMOS.\\ 
& {\color{black} 3.3~mm$^2$} & {\color{black}28~nm CMOS.}\\ 
\hline

Accelerator core & {\color{black}3.0} mW & 27.8~MHz, {\color{black}0.82~V,} 65~nm. \\
power consumption & {\color{black} 1.5 mW} & 27.8~MHz, {\color{black}0.7~V,} 28~nm. \\
\hline

Accelerator core & {\color{black}0.9} $\mu$J & 27.8~MHz, {\color{black} 0.82~V,} 65~nm. \\
energy per & {\color{black} 0.45 $\mu$J} & 27.8~MHz, {\color{black} 0.7~V,} 28~nm. \\

classification {\color{black}(EPC)}. & & \\
\hline

\end{tabular}
\end{center}
\end{table}
\section{Discussions}\label{PaperD_Section: Discussions}


\begin{table*}[hbt!]
\scriptsize
\begin{center}
\caption{{\color{black}Comparison of the ConvCoTM IC accelerator with previous 
works.}} \label{PaperD_Table: Comparison with MNIST ASIC solutions}
\centering

\begin{tabular}{
>{\centering\arraybackslash}p{0.100\textwidth}
|>
{\centering\arraybackslash}p{0.110\textwidth}
|>
{\centering\arraybackslash}p{0.110\textwidth}
|>
{\centering\arraybackslash}p{0.09\textwidth}
|>
{\centering\arraybackslash}p{0.105\textwidth}
|>
{\centering\arraybackslash}p{0.09\textwidth}
}

\hline
& \textbf{This work} 
& \textbf{{\color{black}\textit{Envisaged design. This work scaled to~28~nm{\color{black}.}}}} 
& \textbf{TCAS-I´25 Zhao~\cite{Thesis_10902457_3p32nJ_per_Frame}}
& \textbf{TCAS-II'23 Yejun~\cite{Thesis_10058600_Yejun_Ko_SNN_TCSI_II}}
& \textbf{JSSC'23 \hspace{10pt} Yang~\cite{Thesis_10034979_Yang_TNN_MNIST_May2023}}
\\
\hline
\textbf{Technology} 
& 65 nm CMOS
& 28 nm CMOS
& 28 nm CMOS
& 65 nm CMOS
& 40 nm CMOS 
\\
\hline
\textbf{Active area} 
& 2.7 mm$^2$
& 0.27 mm$^2$
& 0.261 mm$^2$
& 0.57 mm$^2$
& 0.98 mm$^2$
\\
\hline
\textbf{Algorithm} 
& ConvCoTM
& ConvCoTM
& CNN
& SNN
& Ternary CNN 
\\
\hline
\textbf{Design type} 
& Digital
& Digital
& Analog, time domain
& Neuromorphic mixed-signal
& IMC, mixed-signal  
\\
\hline
\textbf{Image dataset} 
& MNIST, FMNIST, KMNIST 
& MNIST, FMNIST, KMNIST
& MNIST
& MNIST
& MNIST 
\\
\hline
\textbf{Measured test accuracy} 
& 97.42\%, 84.54\%, 82.55\%
& 97.42\%, 84.54\%, 82.55\%
& 97.9\%
& 95.35\%
& 97.1\%  
\\
\hline
\textbf{Classifications per second} 
& 60.3~k $^{a}$ \hspace{20pt} 2.27~k $^{b}$ 
& 60.3~k $^{a}$ 
& 3508 
& 233~k~(1.2 V) \hspace{20pt} 40~k~(0.7 V)
& 549 
\\
\hline
\textbf{Latency $^{e}$}
&25.4 $\mu$s $^{a}$ \hspace{20pt} 0.66~ms~$^{b}$ 
& 25.4 $\mu$s $^{a}$ 
& Not stated
& Not stated
& Not stated 
\\
\hline
\textbf{Inference power} 
& 1.15~mW~$^{a,c}$   
0.52~mW~$^{a,d}$ 
81~$\mu$W~$^{b,c}$ 
21~$\mu$W~$^{b,d}$
& 0.26~mW~$^{a,f}$ 
& 11.6 $\mu$W
& 9.37 mW~(1.2 V)  0.517 mW~(0.7 V)
& 96 $\mu$W  
\\
\hline
\textbf{Energy per classification (EPC)} 

& 19.1 nJ~$^{a,c}$
\hspace{20pt} 8.6 nJ~$^{a,d}$
\hspace{20pt} 35.3 nJ~$^{b,c}$
\hspace{20pt} 9.6 nJ~$^{b,d}$
& 4.3 nJ~$^{a,f}$
& 3.32 nJ
& 40.17 nJ~(1.2 V) 12.92 nJ~(0.7 V)
& 0.18 $\mu$J 
\\
\hline
\end{tabular}
\end{center} 
$^a$~27.8~MHz clock frequency. $^b$~1.0~MHz clock frequency. $^c$~The accelerator core supply voltage (vdd) is~1.20~V.  $^d$~vdd=0.82~V. 
$^e$~For processing of a single image sample, including data transfer. 
$^f$~vdd=0.7~V.

\end{table*}

\begin{table*}[hbt!]
\scriptsize
\begin{center}
\caption{{\color{black}Comparison of the envisaged scaled-up ConvCoTM IC accelerator with previous 
works that operate on CIFAR-10.}} \label{PaperD_Table: Comparison with CIFAR10 ASIC solutions}
\centering

\begin{tabular}{
>{\centering\arraybackslash}p{0.120\textwidth}
|>{\centering\arraybackslash}p{0.150\textwidth}
|>{\centering\arraybackslash}p{0.09\textwidth}
|>{\centering\arraybackslash}p{0.10\textwidth}
|>{\centering\arraybackslash}p{0.10\textwidth}
|>{\centering\arraybackslash}p{0.10\textwidth}
|>{\centering\arraybackslash}p{0.10\textwidth}
}

\hline
& \textbf{{\color{black}\textit{Envisaged design from Subsection~\ref{PaperD_subsection_ScaledUp_CIFAR10}}}} 
& \textbf{TCAS-I'20 Mauro~\cite{Thesis_AlwaysOnBNNDiMauro}} 
& \textbf{JSSC'21 \hspace{10pt} Knag~\cite{Thesis_KnagPhilC2021A6AB}} 
& \textbf{TCAS-I'20 Bankman~\cite{Thesis_BankmanBNNmixedsignal}}
& \textbf{{\color{black}TCAS-I'25 Park~\cite{Thesis_10734367_SNN_701p7_TOPSperW}}}
& \textbf{{\color{black}JSSC'25 Yoshioka~\cite{Thesis_ACIM_2025_JSSC_10689660}}}
\\
\hline
\textbf{Technology} 
& 65~nm~CMOS 28~nm~CMOS 
& 22 nm FD-SOI 
& 10 nm \mbox{FinFET} 
& 28 nm CMOS
& 65 nm CMOS
& 65 nm CMOS
\\
\hline
\textbf{Active area} 
& 17.7~mm$^2$~(65~nm) 3.3~mm$^2$~(28~nm)
& 2.3 mm$^2$
& 0.39 mm$^2$
& 4.6 mm$^2$
& 0.17 mm$^2$
& 0.48 mm$^2$
\\
\hline
\textbf{Algorithm} 
& ConvCoTM
& BNN (SW defined)
& BNN
& BNN 
& SNN (spiking VGG-16)
& CNN / Transformer
\\
\hline
\textbf{Design type} 
& Digital
& Digital (SoC)
& Digital
& IMC, mixed-signal  
& Analog, with time-domain IMC
& Analog, IMC
\\
\hline
\textbf{Image dataset} 
& CIFAR-10 
& CIFAR-10 
& CIFAR-10 
& CIFAR-10 
& CIFAR-10 
& CIFAR-10 
\\
\hline
\textbf{Test accuracy} 
& 79\% 
& 99\% of nominal accuracy 
& 86\% 
& 86\% 
& 91.13\%
& 91.7\%~$^{j}$ \hspace{20pt} 95.8\%~$^{k}$
\\
\hline
\textbf{Classifications per second} 
& 3440 $^{a}$ 
& 15.4 
& Not stated
& 237
& Not stated
& Not stated
\\
\hline
\textbf{Latency $^{b}$}
&0.3~ms~$^{a}$ 
& Not stated 
& Not stated 
& Not stated 
& Not stated
& Not stated
\\
\hline
\textbf{Inference power} 
& 3.0~mW~$^{a,c}$~(65~nm)  1.5~mW~$^{a,d}$~(28~nm) 
& 674 $\mu$W~$^{e}$  
& 5.6~mW~$^{l}$ 607~mW~$^{m}$
& 0.9 mW 
& 0.55~mW~$^i$
& Not stated
\\
\hline
\textbf{TOPS/W} 
& Not stated~$^{f}$
& 13~$^{e,n}$ 
&617~$^{l}$\hspace{20pt}269~$^{m}$
& 532
&701.7~$^{g}$\hspace{20pt}488.7~$^{h}$
&4094~$^{j}$\hspace{20pt}818~$^{k}$
\\
\hline
\textbf{Energy per classification (EPC)} 
& 0.9~$\mu$J~$^{a,c}$~(65~nm) 0.45~$\mu$J~$^{a,d}$~(28~nm) 
& 43.8 $\mu$J 
& Not stated
& 3.8 $\mu$J 
& Not stated
& Not stated
\\
\hline
\end{tabular}
\end{center} 
$^a$~27.8~MHz clock frequency. 
$^b$~For processing of a single image sample, including data transfer. 
$^c$~vdd=0.82~V. $^d$~vdd=0.7~V. 
$^e$~For the complete SoC. 
$^f$~Not stated due to the very different architecture compared to DNNs and SNNs.
$^g$~Macro. 
$^h$~System. 
$^i$ 50~MHz, 0.75~V. 
$^j$ CNN. 
$^k$ Transformer. 
$^l$ 13~MHz, 0.37~V. 
$^m$ 622~MHz, 0.75~V. 
$^n$ 0.5~V. 

\end{table*}


We consider the ConvCoTM accelerator with~128 clauses as a small configuration. In light of this, we are satisfied with the obtained test accuracy which is comparable with other low-complexity and ULP solutions. {{\color{black}Table~\ref{PaperD_Table: Comparison with MNIST ASIC solutions} shows the performance and properties of state-of-the-art ULP IC implementations for classification of MNIST. There, we have also included the performance of the envisaged~28~nm CMOS implementation of the ConvCoTM accelerator, described in Subsection~\ref{PaperD_subsec_ASIC_scaled_to_28nm}. In Table~\ref{PaperD_Table: Comparison with CIFAR10 ASIC solutions} the envisaged accelerator design from Subsection~\ref{PaperD_subsection_ScaledUp_CIFAR10} is compared with other ULP accelerators that operate on the CIFAR-10 dataset.}

Inspired by the \textit{TinyML}~\cite{Thesis_banbury2021mlperf} benchmarks for embedded systems, we specify the \textit{classification rate}, the \textit{peak power consumption} and the \textit{EPC}. These are system parameters which are independent of the accelerator model and architecture~\cite{Thesis_9177369_Sze_TOPS_alone}. However, when used for comparing solutions, one must be careful to specify the ML task performed. 
The CIFAR-10 dataset is significantly more challenging than MNIST. Therefore, accelerators that operate on CIFAR-10,
have lower classification rates and higher EPC compared to accelerators that operate on MNIST. 

The mixed-signal IMC-based solution in~\cite{Thesis_10034979_Yang_TNN_MNIST_May2023} operates with a peak power consumption of~96~$\mu$W and consumes~0.18~$\mu$J per classification on MNIST, with a test accuracy of ~97.1\%. 
In~\cite{Thesis_10058600_Yejun_Ko_SNN_TCSI_II} a mixed-signal SNN IC accelerator achieves an EPC of~12.92~nJ with~0.7~V supply voltage. 
It utilizes a time-domain echo state network (ESN)~\cite{Thesis_LUKOSEVICIUS2009127_Echo_State_Network} approach and thus requires non-standard interfacing with the surrounding system's resources. The test accuracy achieved on MNIST is~95.35\%.
The~28~nm CMOS CNN-based accelerator in~\cite{Thesis_10902457_3p32nJ_per_Frame} utilizes an IMC architecture with analog and time-domain signal processing. It achieves an EPC of~3.32~nJ, which is the lowest reported for~MNIST.

Our ConvCoTM accelerator ASIC achieves an EPC of~8.6~nJ at~0.82~V operating voltage. This is the second lowest EPC reported, for a manufactured ASIC that operates on MNIST. 
As the ConvCoTM ASIC is fully digital, the classification rate and thereby also the peak power consumption, can be scaled by adjusting the clock frequency. At a clock frequency of~1~MHz {\color{black}and a supply voltage of 0.82~V, the achieved classification rate was~2.27~k~FPS at a power consumption of~{\color{black}21}~$\mu$W.}

Most ULP accelerators are based on analog techniques~\cite{Thesis_10902457_3p32nJ_per_Frame, Thesis_10058600_Yejun_Ko_SNN_TCSI_II, Thesis_10034979_Yang_TNN_MNIST_May2023, Thesis_BankmanBNNmixedsignal}. The test accuracy of such solutions is susceptible to variations in process parameters, supply voltage and operating temperature~(PVT).
All-digital solutions, like our ConvCoTM ASIC and~\cite{Thesis_AlwaysOnBNNDiMauro, Thesis_KnagPhilC2021A6AB, Thesis_YodaNN7878541}, avoid such non-ideal effects. In addition, these accelerators are easier to integrate in larger SoCs, and can more easily be scaled and re-targeted to other technologies. This applies for CNNs as well as TMs.

In the ConvCoTM accelerator, the combinational clause logic draws only a small amount of energy compared to the clock tree of the inference-core DFFs. This is the main reason why the CSRF technique (see Subsection~\ref{PaperD_subsec: Clause Pool}) did not provide a greater reduction in EPC. 
Other convolutional TM architectures, that operate at higher clock frequencies, process more literals in parallel, and utilize sparsity in TM models, can potentially benefit more from the CSRF technique.

{
\color{black}

The estimated EPC of the envisaged design described in Subsection~\ref{PaperD_subsection_ScaledUp_CIFAR10}, for operation on CIFAR-10 images, is very low. However, the test accuracy achieved with the TM algorithm for multi-channel images, is still not at the same level as for CNNs, BNNs and SNNs. Further research is required to obtain improvement on this performance parameter. The accelerators in~\cite{Thesis_KnagPhilC2021A6AB, Thesis_10734367_SNN_701p7_TOPSperW, Thesis_ACIM_2025_JSSC_10689660} achieve very high performance in terms of \mbox{TOPS/W}. However, as the classification rates are not specified, one cannot easily calculate the EPC numbers for these solutions.}

{\color{black}
An overview of select TM-based HW solutions, including this work, is shown in Table~\ref{PaperD_Table_TM_HW_solutions}.
The solutions span from all-digital to mixed-signal IMC. There are two silicon-proven ASICs and many FPGA solutions. Several TM concepts have been evaluated based on simulations.}


\begin{table*}[hbt!]
\scriptsize
\begin{center}
\caption{{\color{black}Overview of TM-based HW solutions.}} \label{PaperD_Table_TM_HW_solutions}
\centering

\begin{tabular}{
>{\centering\arraybackslash}p{0.08\textwidth}
|>{\centering\arraybackslash}p{0.090\textwidth}
|>{\centering\arraybackslash}p{0.09\textwidth}
|>{\centering\arraybackslash}p{0.08\textwidth}
|>{\centering\arraybackslash}p{0.08\textwidth}
|>{\centering\arraybackslash}p{0.09\textwidth}
|>{\centering\arraybackslash}p{0.09\textwidth}
|>{\centering\arraybackslash}p{0.09\textwidth}
|>{\centering\arraybackslash}p{0.09\textwidth}
}

\hline
& \textbf{This work} 
& \textbf{Phil.~Trans.~A'20 Wheeldon~\cite{Thesis_Wheeldon1}}
& \textbf{TCAS-I'25 Mao~\cite{Thesis_Dynamic_TM_FPGA_10990173}}
& \textbf{TCAS-I'25 Tunheim~\cite{Thesis_10812055_TCSI_ConvCoTM28x28_FPGA}}
& \textbf{ISTM'23 Sahu~\cite{Thesis_10455016_Prajwal}}
& \textbf{MICPRO'23 Tunheim~\cite{Thesis_MICPROTUNHEIM2023104949}}
& \textbf{ISLPED'23 Ghazal~\cite{Thesis_ghazal2023imbue}}
& \textbf{Phil.~Trans.~A'25 Ghazal~\cite{Thesis_ghazal2024inmemorylearningautomataarchitecture}}
\\
\hline
\textbf{Technology and design status} 
& ASIC, 65 nm CMOS, silicon proven
& ASIC, 65 nm CMOS, silicon-proven
& FPGA
& FPGA
& FPGA
& FPGA
& ASIC, simulation 
& ASIC, simulation
\\
\hline
\textbf{Algorithm} 
& ConvCoTM
& Vanilla TM
& Vanilla TM, CoTM
& ConvCoTM
& Vanilla TM
& CTM
& Vanilla TM
& CoTM
\\
\hline
\textbf{Operation} 
& Inference
& Inference
& Training and inference
& Training and inference
& Inference
& Training and Inference
& Inference
& Inference
\\
\hline
\textbf{Design type} 
& Digital
& Digital
& Digital
& Digital
& Digital
& Digital
& IMC (ReRAM), mixed-signal
& IMC (Y-flash), mixed-signal
\\
\hline
\textbf{Dataset} 
& MNIST, FMNIST, KMNIST 
& Binary IRIS
& MNIST, FMNIST, KMNIST
& MNIST, FMNIST, KMNIST
& MNIST
& 2D Noisy \hspace{10pt} XOR~$^c$
& MNIST, FMNIST, KMNIST, KWS-6~$^d$
& MNIST

\\
\hline
\textbf{Test accuracy} 
& 97.42\%, 84.54\%, 82.55\%
& 97.0\%
& 97.74\%, 86.38\%, 83.11\%
& 97.6\%, \hspace{4pt}84.1\%, \hspace{4pt}82.8\%
& 97.71\%
& 99.9\% 
& 96.48\%, 87.67\%, 88.6\%, 87.1\%
& 96.3\%
\\
\hline
\textbf{Classifications per second} 
& 60.3~k $^{a}$ 
& Not stated
& 22.4~k
& 134~k
& Not stated~$^{f}$
& 4.4~M
& Not stated
& Not stated
\\
\hline
\textbf{Inference power} 
& 0.52~mW~$^{a,b}$   
& Not stated
& 1.65~W
& 1.8~W 
& Not stated
& 2.529~W
& Not stated
& Not stated
 
\\
\hline
\textbf{Energy per classification}
& 8.6 nJ~$^{a,b}$
& Not stated~$^e$
& 73.6~$\mu$J
& 13.3~$\mu$J 
& Not stated
& 0.6~$\mu$J
& 13.9~nJ, 23.66~nJ, 26.47~nJ, 5.91~nJ 
& Not stated~$^g$
\\
\hline
\end{tabular}
\end{center} 
$^a$~27.8~MHz clock frequency. 
$^b$~vdd=0.82~V 
$^c$~$4 \times 4$ booleanized images. 
$^d$~Keyword spotting dataset with six words.
$^e$~62.7 TOP/J. 
$^f$~The number of clock cycles required per classification is stated, for different number of clauses and with/without literal budgeting.
$^g$~24.56 TOPS/W.


\end{table*}

\section{Conclusion}\label{PaperD_Conclusion}

In this work, an ASIC implementation of a ConvCoTM accelerator for inference is presented. It operates on booleanized images of~28$\times$28 pixels, and samples are classified into~10 categories. A pool of~128 clauses, i.e., pattern recognition logical expressions, is applied for the accelerator. The solution is fully programmable, and achieves a  test accuracy of~97.42\% on MNIST. With a clock frequency of~27.8~MHz, the accelerator's throughput is~60.3~k images per second, and with an operating voltage of~0.82~V, the energy consumed by the accelerator core per classification is~8.6~nJ. We believe the TM represents an attractive ML solution for low-power edge nodes in IoT systems for various workloads. The ASIC demonstrates the simplicity, hardware-friendliness and power-efficiency of \mbox{all-digital} TM solutions.
\section*{Acknowledgments}

The authors would like to thank Adrian Wheeldon (Literal Labs, UK) and Jordan Morris (UK) for valuable general ASIC implementation discussions. Thank you also to Joel Trickey (STFC, UK) for support with IC design backend tools, and to Alberto Pagotto (imec.IC-Link, Belgium) for help during the ASIC tapeout phase. Finally, we would like to thank Geir Rune Angell (Westcontrol, Norway) for the design of the ASIC test board. 

\bibliographystyle{IEEEtran}
\bibliography{References_ASIC_updated}

\begin{thebibliography}{10}
\providecommand{\url}[1]{#1}
\csname url@samestyle\endcsname
\providecommand{\newblock}{\relax}
\providecommand{\bibinfo}[2]{#2}
\providecommand{\BIBentrySTDinterwordspacing}{\spaceskip=0pt\relax}
\providecommand{\BIBentryALTinterwordstretchfactor}{4}
\providecommand{\BIBentryALTinterwordspacing}{\spaceskip=\fontdimen2\font plus
\BIBentryALTinterwordstretchfactor\fontdimen3\font minus \fontdimen4\font\relax}
\providecommand{\BIBforeignlanguage}[2]{{%
\expandafter\ifx\csname l@#1\endcsname\relax
\typeout{** WARNING: IEEEtran.bst: No hyphenation pattern has been}%
\typeout{** loaded for the language `#1'. Using the pattern for}%
\typeout{** the default language instead.}%
\else
\language=\csname l@#1\endcsname
\fi
#2}}
\providecommand{\BIBdecl}{\relax}
\BIBdecl

\bibitem{Thesis_maheepala2020low}
M.~Maheepala, M.~A. Joordens, and A.~Z. Kouzani, ``Low power processors and image sensors for vision-based {IoT} devices: {A} review,'' \emph{IEEE Sensors Journal}, vol.~21, no.~2, pp. 1172--1186, 2020.

\bibitem{Thesis_8662396LeCunISSCC2019}
Y.~LeCun, ``{D}eep learning hardware: {P}ast, present, and future,'' in \emph{2019 IEEE International Solid-State Circuits Conference - (ISSCC)}, 2019, pp. 12--19.

\bibitem{Thesis_8114708_Sze_survey_IEE_Proceedings}
V.~Sze, Y.-H. Chen, T.-J. Yang, and J.~S. Emer, ``Efficient processing of deep neural networks: {A} tutorial and survey,'' \emph{Proceedings of the IEEE}, vol. 105, no.~12, pp. 2295--2329, 2017.

\bibitem{Thesis_9063049_Jeff_dean}
J.~Dean, ``The deep learning revolution and its implications for computer architecture and chip design,'' in \emph{IEEE ISSCC}, 2020, pp. 8--14.

\bibitem{Thesis_BankmanBNNmixedsignal}
D.~Bankman, L.~Yang, B.~Moons, M.~Verhelst, and B.~Murmann, ``An always-on 3.8 $\mu${J}/86\% {CIFAR-10} mixed-signal binary {CNN} processor with all memory on chip in 28-nm {CMOS},'' \emph{IEEE JSSC}, vol.~54, pp. 158--172, 1 2019.

\bibitem{Thesis_AlwaysOnBNNDiMauro}
A.~D. Mauro, F.~Conti, P.~D. Schiavone, D.~Rossi, and L.~Benini, ``Always-on 674$\mu${W@4GOP/s} error resilient binary neural networks with aggressive {SRAM} voltage scaling on a 22-nm {IoT} end-node,'' \emph{IEEE TCAS-I}, vol.~67, no.~11, pp. 3905--3918, 2020.

\bibitem{Thesis_KnagPhilC2021A6AB}
P.~C. Knag, G.~K. Chen, H.~E. Sumbul, R.~Kumar, S.~K. Hsu, A.~Agarwal, M.~Kar, S.~Kim, M.~A. Anders, H.~Kaul, and R.~K. Krishnamurthy, ``\BIBforeignlanguage{eng}{A 617-{TOPS}/{W} all-digital binary neural network accelerator in 10-nm {FinFET} {CMOS}},'' \emph{\BIBforeignlanguage{eng}{IEEE JSSC}}, vol.~56, no.~4, pp. 1082--1092, 2021.

\bibitem{Thesis_YodaNN7878541}
R.~Andri, L.~Cavigelli, D.~Rossi, and L.~Benini, ``{YodaNN}: {A}n architecture for ultralow power binary-weight {CNN} acceleration,'' \emph{IEEE TCAD}, vol.~37, pp. 48--60, 1 2018.

\bibitem{Thesis_10034979_Yang_TNN_MNIST_May2023}
X.~Yang, K.~Zhu, X.~Tang, M.~Wang, M.~Zhan, N.~Lu, J.~P. Kulkarni, D.~Z. Pan, Y.~Liu, and N.~Sun, ``An in-memory-computing charge-domain ternary {CNN} classifier,'' \emph{IEEE JSSC}, vol.~58, no.~5, pp. 1450--1461, 2023.

\bibitem{Thesis_OrigTM}
O.-C. Granmo, ``The {T}setlin machine -- a game theoretic bandit driven approach to optimal pattern recognition with propositional logic,'' 2018, arXiv:1804.01508.

\bibitem{Thesis_Wheeldon1}
A.~Wheeldon, R.~Shafik, T.~Rahman, J.~Lei, A.~Yakovlev, and O.-C. Granmo, ``Learning automata based energy-efficient {AI} hardware design for {I}o{T} applications,'' \emph{Philosophical Transactions of the Royal Society A: Mathematical, Physical and Engineering Sciences}, vol. 378, no. 2182, 2020.

\bibitem{Thesis_10812055_TCSI_ConvCoTM28x28_FPGA}
S.~A. Tunheim, L.~Jiao, R.~Shafik, A.~Yakovlev, and O.-C. Granmo, ``{T}setlin machine-based image classification {FPGA} accelerator with on-device training,'' \emph{IEEE Transactions on Circuits and Systems I: Regular Papers}, vol.~72, no.~2, pp. 830--843, 2025.

\bibitem{Thesis_CTM}
O.-C. Granmo, S.~Glimsdal, L.~Jiao, M.~Goodwin, C.~W. Omlin, and G.~T. Berge, ``The convolutional {T}setlin machine,'' 2019, arXiv: 1905.09688.

\bibitem{Thesis_lecun-mnisthandwrittendigit-2010}
\BIBentryALTinterwordspacing
Y.~LeCun and C.~Cortes, ``{MNIST} handwritten digit database,'' 2010. [Online]. Available: \url{http://yann.lecun.com/exdb/mnist/}
\BIBentrySTDinterwordspacing

\bibitem{Thesis_Fashion-MNIST_dataset}
{Fashion-MNIST Repo.}, \url{https://www.openml.org/search?type=data&status=active&id=40996&sort=runs}, 2017.

\bibitem{Thesis_Kuzushiji-MNIST_dataset}
{Kuzushiji-MNIST Repo.}, \url{https://github.com/rois-codh/kmnist?tab=readme-ov-file}, 2019.

\bibitem{Thesis_granmo2023tmcomposites}
O.-C. Granmo, ``{TMC}omposites: {P}lug-and-play collaboration between specialized {T}setlin machines,'' 2023, arXiv: 2309.04801.

\bibitem{Thesis_grønningsæter2024optimizedtoolboxadvancedimage}
Y.~Grønningsæter, H.~S. Smørvik, and O.-C. Granmo, ``An optimized toolbox for advanced image processing with {T}setlin machine composites,'' in \emph{2024 International Symposium on the {T}setlin Machine ({ISTM})}, 2024, pp. 1--8.

\bibitem{Thesis_CoalescedTM}
S.~Glimsdal and O.-C. Granmo, ``Coalesced multi-output {T}setlin machines with clause sharing,'' 2021, arXiv:2108.07594.

\bibitem{Thesis_10902457_3p32nJ_per_Frame}
Y.~Zhao, P.~He, Y.~Zhu, R.~P. Martins, C.-H. Chan, and M.~Zhang, ``A 28-nm 3.32-n{J}/frame compute-in-memory {CNN} processor with layer fusion for always-on applications,'' \emph{IEEE Transactions on Circuits and Systems I: Regular Papers}, pp. 1--13, 2025.

\bibitem{Thesis_10058600_Yejun_Ko_SNN_TCSI_II}
Y.~Ko, S.~Kim, K.~Shin, Y.~Park, S.~Kim, and D.~Jeon, ``A 65 nm 12.92-n{J}/inference mixed-signal neuromorphic processor for image classification,'' \emph{IEEE Transactions on Circuits and Systems II: Express Briefs}, vol.~70, no.~8, pp. 2804--2808, 2023.

\bibitem{Thesis_XNORNeuralEngine}
F.~Conti, P.~Schiavone, and L.~Benini, ``{XNOR} neural engine: {A} hardware accelerator {IP} for 21.6 f{J}/op binary neural network inference,'' \emph{IEEE TCAD}, vol.~37, pp. 2940--2951, 11 2018.

\bibitem{Thesis_Krizhevsky09learningmultiple}
A.~Krizhevsky, ``Learning multiple layers of features from tiny images,'' \url{https://www.cs.toronto.edu/~kriz/cifar.html}, 2009.

\bibitem{Thesis_10396030_Yuchao_Zhang_SNN_ASICON}
Y.~Zhang, Z.~Xuan, and Y.~Kang, ``A 28nm 15.09n{J}/inference neuromorphic processor with {SRAM}-based charge domain in-memory-computing,'' in \emph{2023 IEEE 15th International Conference on ASIC (ASICON)}, 2023, pp. 1--4.

\bibitem{Thesis_10388821_Wenbing_Fang_SNN_BioCAS}
W.~Fang, Z.~Xuan, S.~Chen, and Y.~Kang, ``An 1.38n{J}/inference clock-free mixed-signal neuromorphic architecture using {ReL-PSP} function and computing-in-memory,'' in \emph{2023 IEEE Biomedical Circuits and Systems Conference (BioCAS)}, 2023, pp. 1--5.

\bibitem{Thesis_10734367_SNN_701p7_TOPSperW}
K.~Park, H.~Jeong, S.~Kim, J.~Shin, M.~Kim, and K.~Jason~Lee, ``A 701.7 {TOPS/W} compute-in-memory processor with time-domain computing for spiking neural network,'' \emph{IEEE Transactions on Circuits and Systems I: Regular Papers}, vol.~72, no.~1, pp. 25--35, 2025.

\bibitem{Thesis_ACIM_2025_JSSC_10689660}
K.~Yoshioka, ``A 818–4094 {TOPS/W} capacitor-reconfigured analog {CIM} for unified acceleration of {CNNs} and transformers,'' \emph{IEEE Journal of Solid-State Circuits}, vol.~60, no.~5, pp. 1844--1855, 2025.

\bibitem{Thesis_MICPROTUNHEIM2023104949}
S.~A. Tunheim, L.~Jiao, R.~Shafik, A.~Yakovlev, and O.-C. Granmo, ``Convolutional {T}setlin machine-based training and inference accelerator for {2-D} pattern classification,'' \emph{Microprocessors and Microsystems}, vol. 103, p. 104949, 2023.

\bibitem{Thesis_10455016_Prajwal}
P.~K. Sahu, S.~Boppu, R.~Shafik, S.~A. Tunheim, O.-C. Granmo, and L.~R. Cenkeramaddi, ``Enhancing inference performance through include only literal incorporation in {T}setlin machine,'' in \emph{2023 International Symposium on the Tsetlin Machine (ISTM)}, 2023, pp. 1--8.

\bibitem{Thesis_GitHub_FPGA_repository}
{GitHub repository for a ConvCoTM FPGA solution with on-device training}, \url{https://github.com/satunheim/ConvCoTM-FPGA-28x28}.

\bibitem{Thesis_Dynamic_TM_FPGA_10990173}
G.~Mao, T.~Rahman, S.~Maheshwari, B.~Pattison, Z.~Shao, R.~Shafik, and A.~Yakovlev, ``Dynamic {T}setlin machine accelerators for on-chip training using {FPGAs},'' \emph{IEEE Transactions on Circuits and Systems I: Regular Papers}, pp. 1--14, 2025.

\bibitem{Thesis_9474126_asynch}
A.~Wheeldon, A.~Yakovlev, R.~Shafik, and J.~Morris, ``Low-latency asynchronous logic design for inference at the edge,'' in \emph{2021 Design, Automation \& Test in Europe Conference \& Exhibition (DATE)}, 2021, pp. 370--373.

\bibitem{Thesis_9565438_asynch}
A.~Wheeldon, A.~Yakovlev, and R.~Shafik, ``Self-timed reinforcement learning using {T}setlin machine,'' in \emph{2021 27th IEEE International Symposium on Asynchronous Circuits and Systems (ASYNC)}, 2021, pp. 40--47.

\bibitem{Thesis_10666312_asynch}
T.~Lan, O.~Ghazal, S.~Ojukwu, K.~Krishnamurthy, R.~Shafik, and A.~Yakovlev, ``An asynchronous winner-takes-all arbitration architecture for {T}setlin machine acceleration,'' in \emph{2024 22nd IEEE Interregional NEWCAS Conference (NEWCAS)}, 2024, pp. 16--20.

\bibitem{Thesis_ghazal2023imbue}
O.~Ghazal, S.~Singh, T.~Rahman, S.~Yu, Y.~Zheng, D.~Balsamo, S.~Patkar, F.~Merchant, F.~Xia, A.~Yakovlev \emph{et~al.}, ``{IMBUE:} {I}n-memory boolean-to-current inference architecture for {T}setlin machines,'' in \emph{2023 IEEE/ACM International Symposium on Low Power Electronics and Design (ISLPED)}.\hskip 1em plus 0.5em minus 0.4em\relax IEEE, 2023, pp. 1--6.

\bibitem{Thesis_ghazal2024inmemorylearningautomataarchitecture}
O.~Ghazal, W.~Wang, S.~Kvatinsky, F.~Merchant, A.~Yakovlev, and R.~Shafik, ``{IMPACT:} {I}n-memory computing architecture based on {Y}-flash technology for coalesced {T}setlin machine inference,'' \emph{Philosophical Transactions of the Royal Society A: Mathematical, Physical and Engineering Sciences}, vol. 383, no. 2288, 2025.

\bibitem{Thesis_Tsetlin1961}
M.~L. Tsetlin, ``On behaviour of finite automata in random medium.'' \emph{Avtomat. i Telemekh, 22(10)}, pp. 1345--1354, 1961.

\bibitem{Thesis_buckman2018thermometer}
J.~Buckman, A.~Roy, C.~Raffel, and I.~Goodfellow, ``Thermometer encoding: {O}ne hot way to resist adversarial examples,'' in \emph{ICLR}, 2018.

\bibitem{Thesis_AXI4StreamProtocol}
{AMBA AXI4 interface protocol}, \url{https://www.xilinx.com/products/intellectual-property/axi.html }.

\bibitem{Thesis_GitHub_ASIC_repository}
{GitHub repository for a ConvCoTM inference ASIC}, \url{https://github.com/satunheim/ConvCoTM_Inference_Accelerator}.

\bibitem{Thesis_TMU_CoalescedClassifier}
{Tsetlin machine unified {(TMU)} {CoTM} github repository}, \url{https://github.com/cair/tmu/blob/main/tmu/models/classification/coalesced_classifier.py}, 2023.

\bibitem{Thesis_ijcai2023p378_constraining_clause_size}
\BIBentryALTinterwordspacing
K.~D. Abeyrathna, A.~A.~O. Abouzeid, B.~Bhattarai, C.~Giri, S.~Glimsdal, O.-C. Granmo, L.~Jiao, R.~Saha, J.~Sharma, S.~A. Tunheim, and X.~Zhang, ``Building concise logical patterns by constraining {T}setlin machine clause size,'' in \emph{Proceedings of the Thirty-Second International Joint Conference on Artificial Intelligence, {IJCAI-23}}, E.~Elkind, Ed.\hskip 1em plus 0.5em minus 0.4em\relax International Joint Conferences on Artificial Intelligence Organization, 8 2023, pp. 3395--3403, main Track. [Online]. Available: \url{https://doi.org/10.24963/ijcai.2023/378}
\BIBentrySTDinterwordspacing

\bibitem{Thesis_782564_Dennard_scaling_Borkar}
S.~Borkar, ``Design challenges of technology scaling,'' \emph{IEEE Micro}, vol.~19, no.~4, pp. 23--29, 1999.

\bibitem{Thesis_knuth97}
D.~E. Knuth, \emph{The Art of Computer Programming, Volume 2: {S}eminumerical Algorithms}, 3rd~ed.\hskip 1em plus 0.5em minus 0.4em\relax Boston: Addison-Wesley, 1997.

\bibitem{Thesis_8746699_line_buffer}
H.~Wang, T.~Wang, L.~Liu, H.~Sun, and N.~Zheng, ``Efficient compression-based line buffer design for image/video processing circuits,'' \emph{IEEE Transactions on Very Large Scale Integration (VLSI) Systems}, vol.~27, no.~10, pp. 2423--2433, 2019.

\bibitem{Thesis_banbury2021mlperf}
C.~Banbury, V.~J. Reddi, P.~Torelli, J.~Holleman, N.~Jeffries, C.~Kiraly, P.~Montino, D.~Kanter, S.~Ahmed, D.~Pau, U.~Thakker, A.~Torrini, P.~Warden, J.~Cordaro, G.~D. Guglielmo, J.~Duarte, S.~Gibellini, V.~Parekh, H.~Tran, N.~Tran, N.~Wenxu, and X.~Xuesong, ``{MLPerf} tiny benchmark,'' 2021, arXiv:2106.07597.

\bibitem{Thesis_9177369_Sze_TOPS_alone}
V.~Sze, Y.-H. Chen, T.-J. Yang, and J.~S. Emer, ``How to evaluate deep neural network processors: {TOPS/W} (alone) considered harmful,'' \emph{IEEE Solid-State Circuits Magazine}, vol.~12, no.~3, pp. 28--41, 2020.

\bibitem{Thesis_LUKOSEVICIUS2009127_Echo_State_Network}
M.~Lukoševičius and H.~Jaeger, ``Reservoir computing approaches to recurrent neural network training,'' \emph{Computer Science Review}, vol.~3, no.~3, pp. 127--149, 2009.

\end{thebibliography}


 



\begin{IEEEbiography}
[{\includegraphics[width=1in,height=1.25in,clip,keepaspectratio]{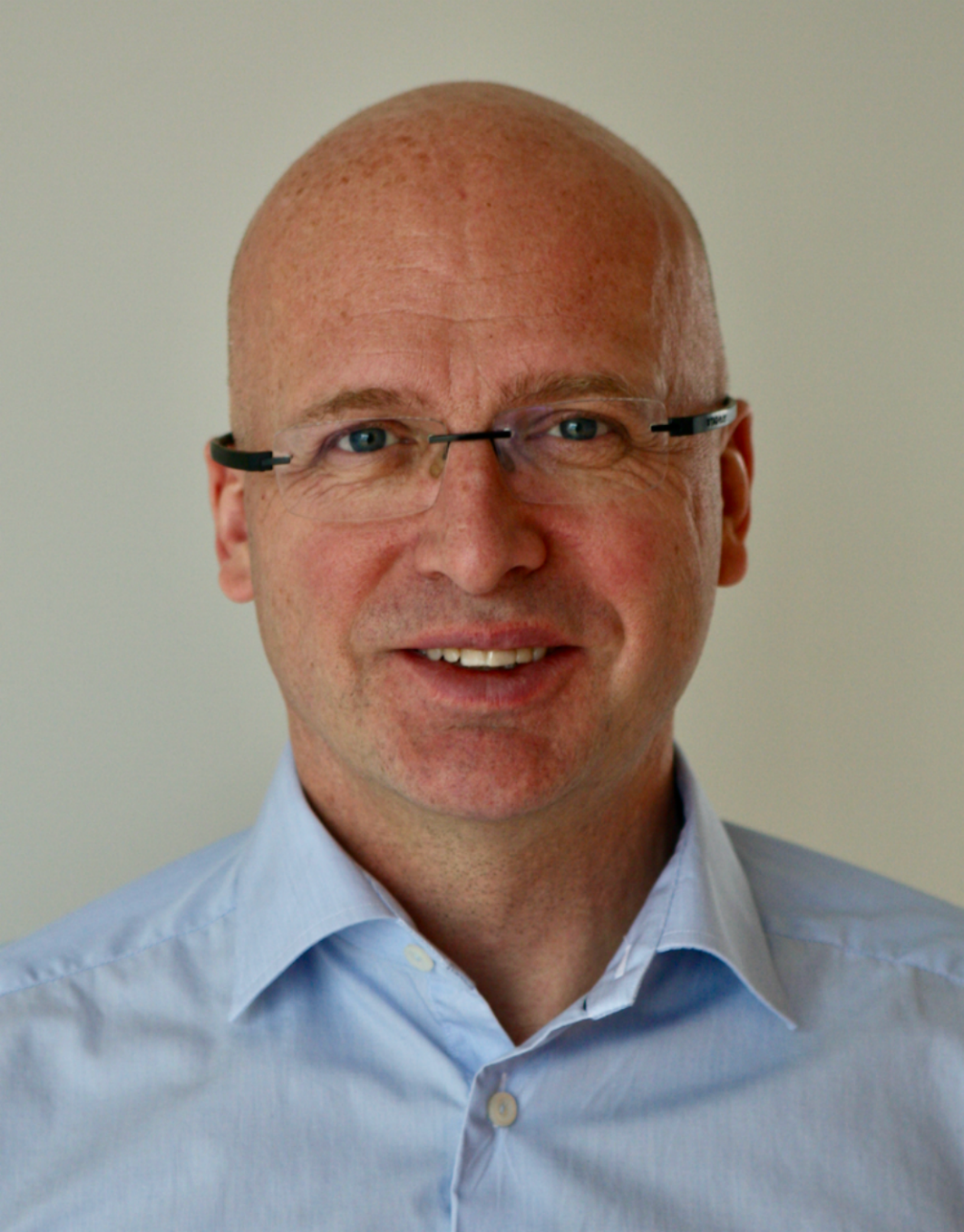}}]{Svein Anders Tunheim} is a PhD research fellow at the University of Agder at Centre for Artificial Intelligence Research (CAIR). He completed his MSc degree in Electrical Engineering at The Norwegian University of Science and Technology (NTNU) in 1991. From 1992 to 1996 he worked as research scientist at SI (Senter for Industriforskning) and SINTEF within the field of mixed-signal integrated circuits (ICs). He was co-founder and Chief Technology Officer at Chipcon, a global supplier of low-power radio frequency ICs and radio protocols. From 2006 to 2008 he worked at Texas Instruments Norway as Technical Director for the Low Power Wireless product line. Currently, at CAIR, he is working on low-power hardware implementations of machine learning systems based on the Tsetlin Machine. He is a Senior Member of IEEE.
\end{IEEEbiography}

\begin{IEEEbiography}
[{\includegraphics[width=1in,height=1.25in,clip,keepaspectratio]{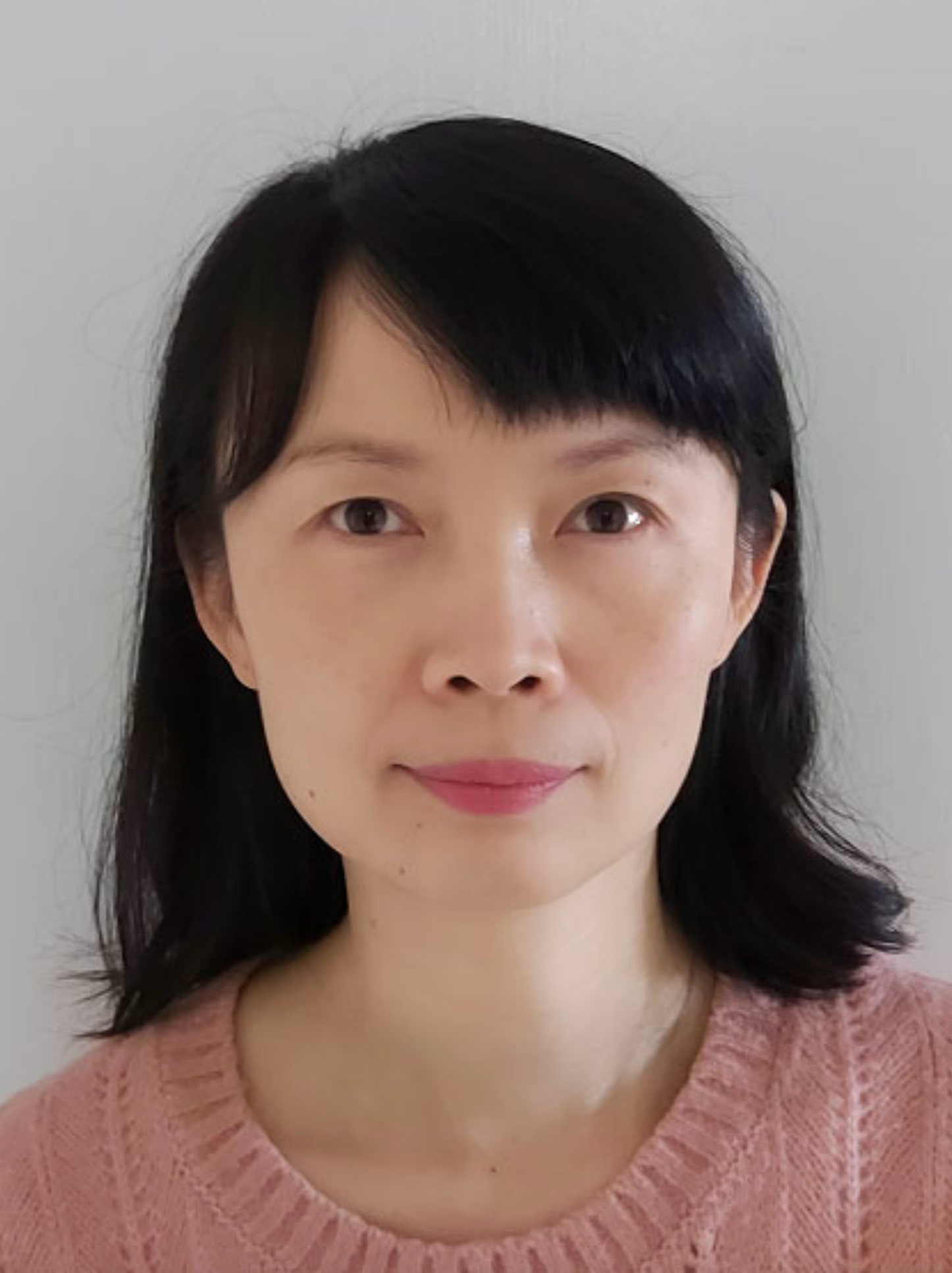}}]{Yujin Zheng} is a PhD candidate in the Microsystems Research Group at Newcastle University. She received her Master's degree in Microelectronics at Newcastle University and a Bachelor's in Applied Electronics from the University of Electronic Science and Technology of China (UESTC). Yujin was a senior board-level hardware engineer in China. She is currently working on ASIC implementation for hardware security.
\end{IEEEbiography}

\begin{IEEEbiography}[{\includegraphics[width=1in,height=1.25in,clip,keepaspectratio]{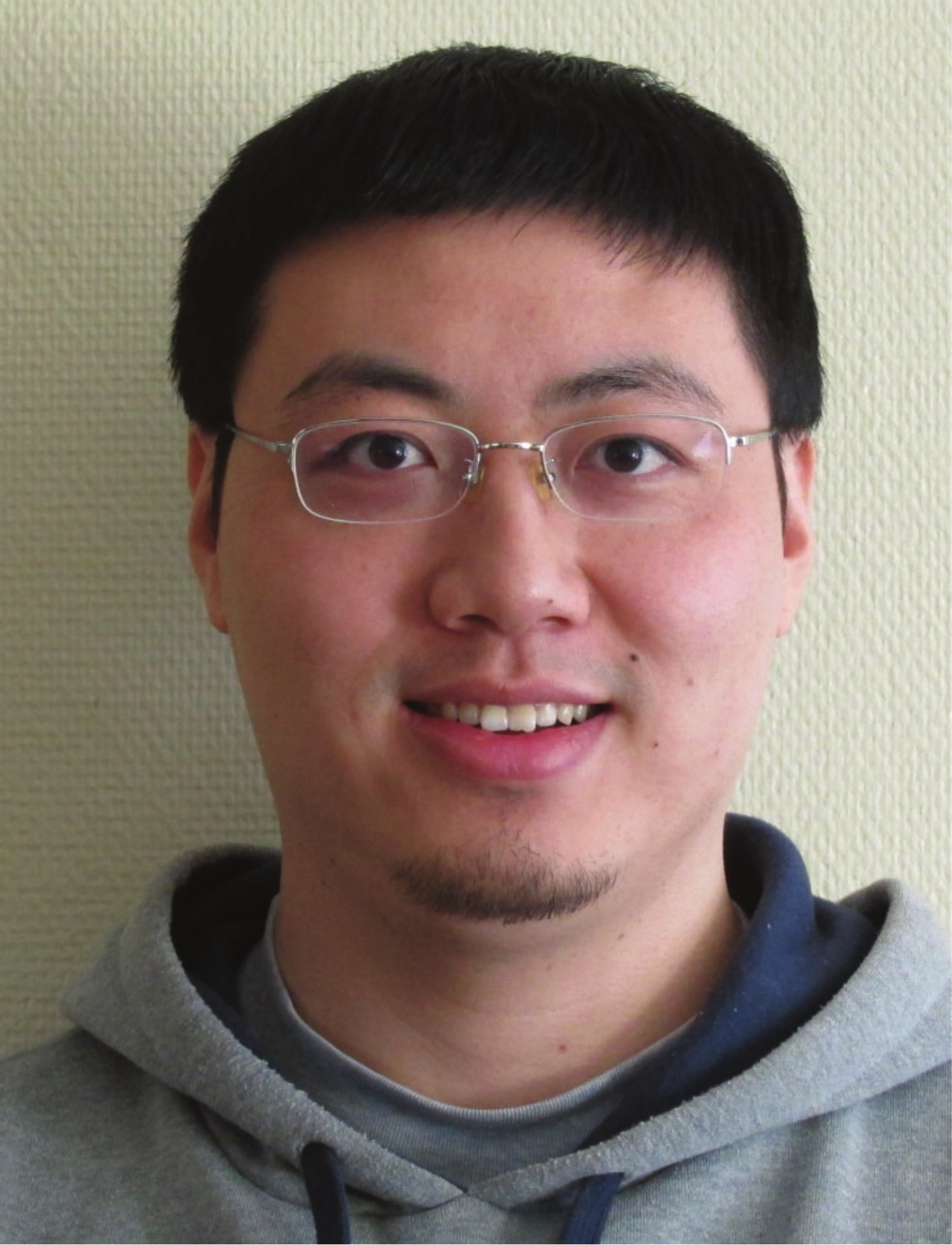}}]{Lei Jiao} received the B.E. degree in telecommunications engineering from Hunan University, Changsha, China, in 2005, the M.E. degree in communication and information system from Shandong University, Jinan, China, in 2008, and the Ph.D. degree in information and communication technology from the University of Agder (UiA), Norway, in 2012. He is currently working as a Professor with the Department of Information and Communication Technology, UiA. His research interests include reinforcement learning, Tsetlin machine, resource allocation and performance evaluation for communication and energy systems. 
\end{IEEEbiography}

\begin{IEEEbiography}
[{\includegraphics[width=1in,height=1.25in,clip,keepaspectratio]{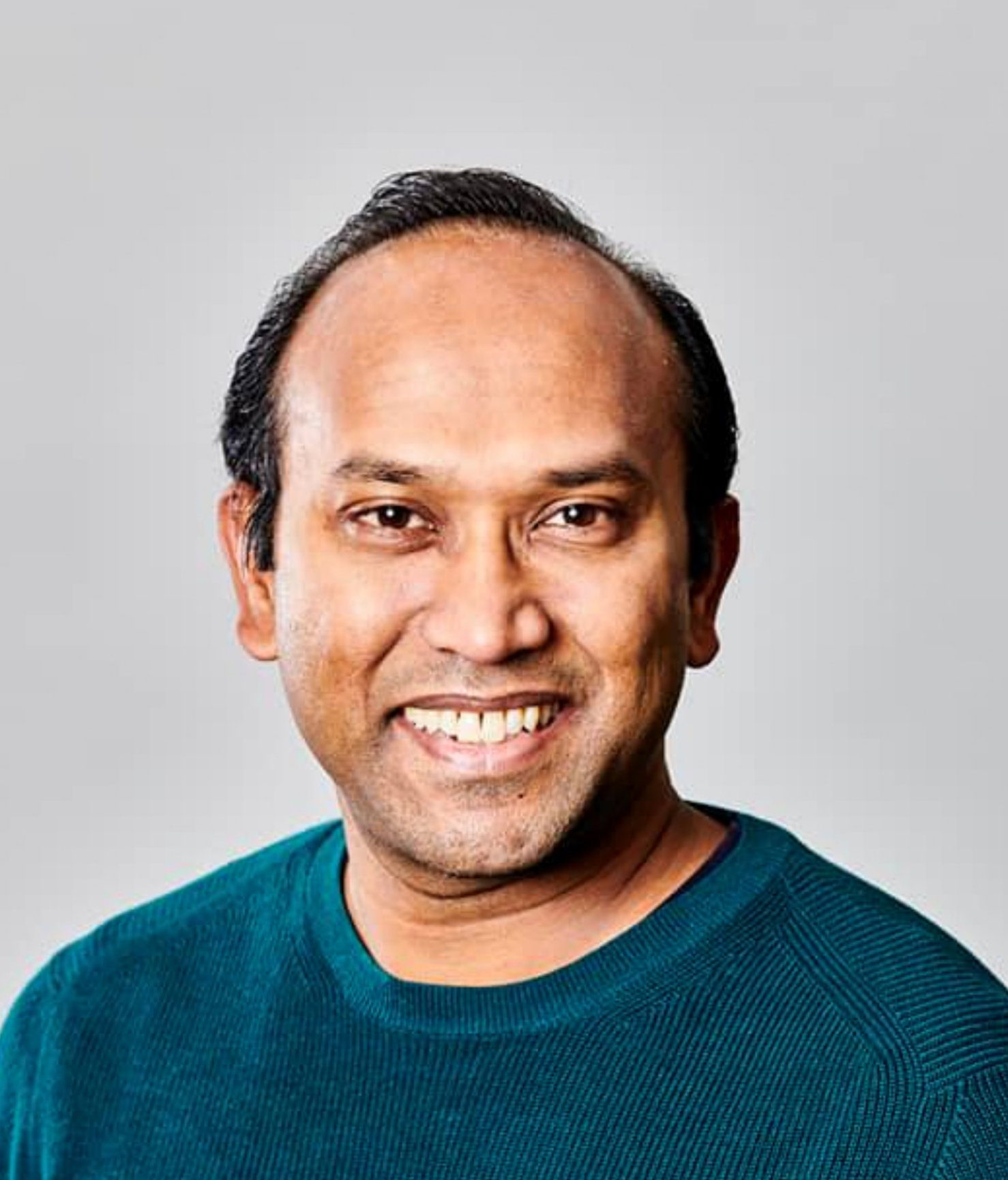}}]{Rishad Shafik} is a Professor in Electronic Systems within the School of Engineering, Newcastle University, UK. Professor Shafik received his PhD, and MSc (with distinction) degrees from Southampton in 2010, and 2005; and BSc (with distinction) from the IUT, Bangladesh in 2001. He is one of the editors of the Springer USA book ``Energy-efficient Fault-tolerant Systems''. He is also author/co-author of 200+ IEEE/ACM peer-reviewed articles, with 4 best paper nominations and 3 best paper/poster awards. He recently chaired multiple international conferences/symposiums, UKCAS2020, ISCAS2025, ISTM2022; guest edited a special theme issue in Royal Society Philosophical Transactions A; he is currently chairing 2nd IEEE SAS, 2025. His research interests include hardware\slash software co-design for energy-efficiency and autonomy.
\end{IEEEbiography}

\begin{IEEEbiography}
[{\includegraphics[width=1in,height=1.25in,clip,keepaspectratio]{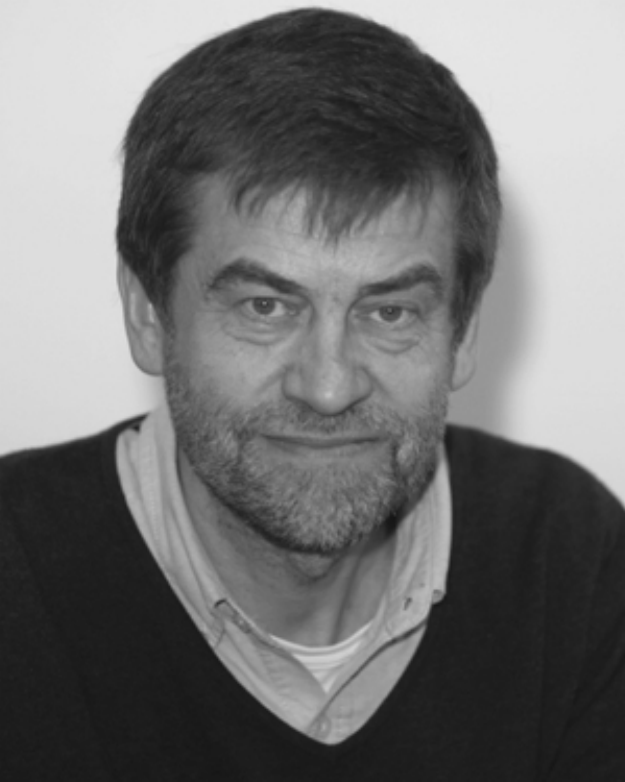}}]{Alex Yakovlev} 
received the Ph.D. degree from the St. Petersburg Electrical Engineering Institute, St. Petersburg, USSR, in 1982, and D.Sc. from Newcastle University, UK, in 2006. He is currently a Professor of Computer Systems Design, who founded and leads the Microsystems Research Group, and co-founded the Asynchronous Systems Laboratory, Newcastle University. He was awarded an EPSRC Dream Fellowship from 2011 to 2013. He has published more than 500 articles in various journals and conferences, in the area of concurrent and asynchronous systems, with several best paper awards and nominations. He has chaired organizational committees of major international conferences. He has been principal investigator on more than 30 research grants and supervised over 70 Ph.D. students. He is a fellow of the Royal Academy of Engineering, UK.
\end{IEEEbiography}

\begin{IEEEbiography}
[{\includegraphics[width=1in,height=1.25in,clip,keepaspectratio]{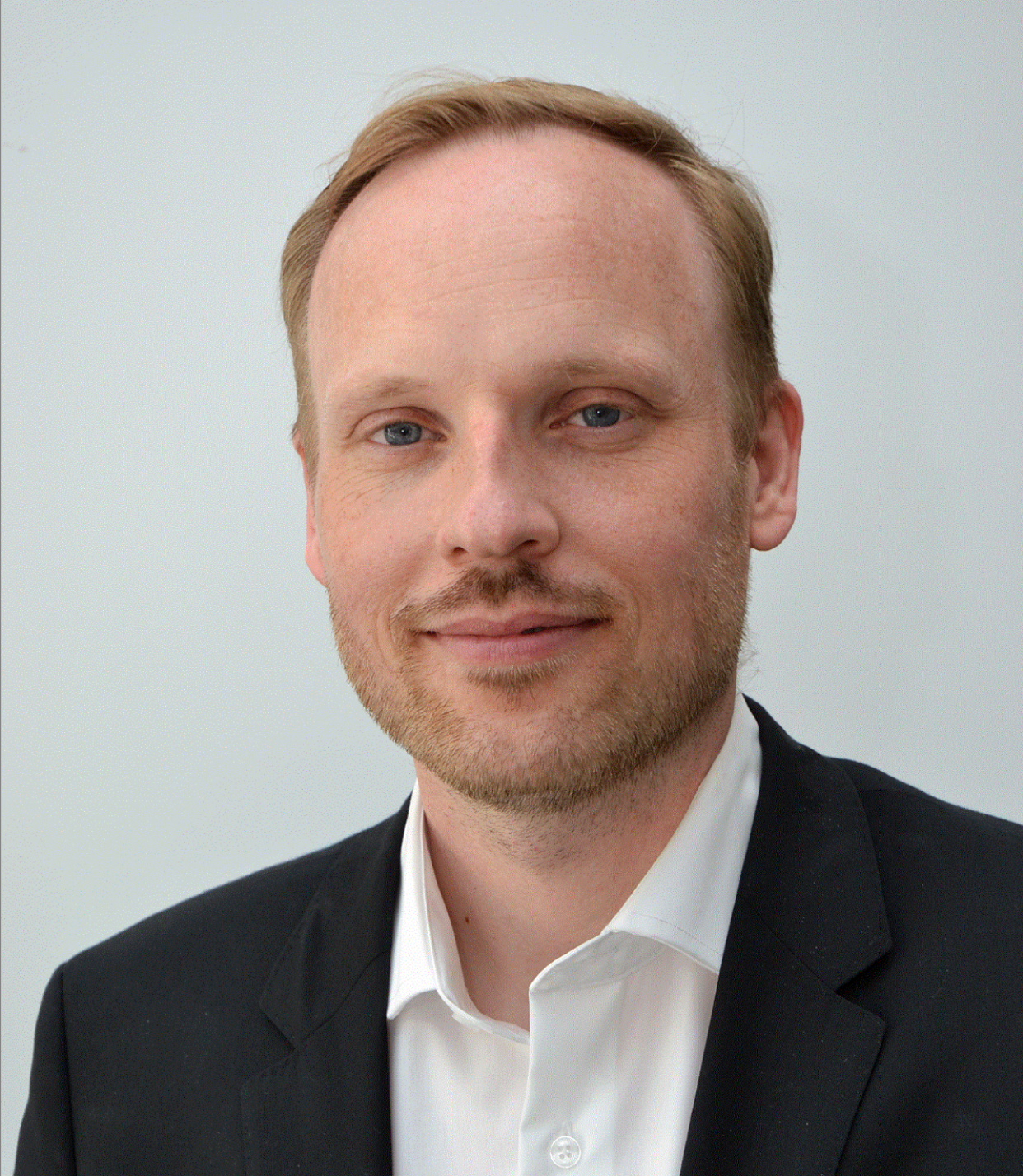}}]{Ole-Christoffer Granmo} is the Founding Director of the Centre for Artificial Intelligence Research (CAIR), University of Agder, Norway. He obtained his master’s degree in 1999 and the PhD degree in 2004, both from the University of Oslo, Norway. In 2018 he created the Tsetlin machine, for which he was awarded the AI researcher of the decade by the Norwegian Artificial Intelligence Consortium (NORA) in 2022. Dr. Granmo has authored more than 160 refereed papers with eight paper awards within machine learning, encompassing learning automata, bandit algorithms, Tsetlin machines, Bayesian reasoning, reinforcement learning, and computational linguistics. He has further coordinated 7+ research projects and graduated 55+ master- and nine PhD students. Dr. Granmo is also a co-founder of NORA. Apart from his academic endeavours, he co-founded the companies Anzyz Technologies AS and Tsense Intelligent Healthcare AS. 
\end{IEEEbiography}






\end{document}